\documentclass{article}

\usepackage{arxiv}

\usepackage[utf8]{inputenc} 
\usepackage[T1]{fontenc}    
\usepackage{hyperref}       
\usepackage{url}            
\usepackage{booktabs}       
\usepackage{amsfonts}       
\usepackage{nicefrac}       
\usepackage{microtype}      
\usepackage{lipsum}
\usepackage{graphicx}
\usepackage{amsmath}
\usepackage{algorithm}
\usepackage{algpseudocode}
\usepackage{multirow}
\usepackage{graphicx}
\usepackage{subcaption}
\usepackage[normalem]{ulem}
\useunder{\uline}{\ul}{}
\graphicspath{ {./images/} }
\usepackage{subcaption}
\usepackage{listings}
\usepackage{xcolor} 
\usepackage{url}
\usepackage{float}

\lstdefinestyle{mystyle}{
    backgroundcolor=\color{white},   
    basicstyle=\ttfamily\small,
    breaklines=true,
    frame=single,
    captionpos=b,
    xleftmargin=\parindent,
    numbers=left,
    numberstyle=\tiny\color{gray},
    keywordstyle=\color{blue},
    commentstyle=\color{green},
    stringstyle=\color{red},
}

\title{DLBacktrace: A Model Agnostic Explainability for any Deep Learning Models}

\author{
  Vinay Kumar Sankarapu\thanks{Equal contribution}, Chintan Chitroda\footnotemark[1], Yashwardhan Rathore\footnotemark[1] \\
  \texttt{\{v.k, chintan.chitroda, yashwardhan.rathore\}@aryaxai.com} \\
\And
  Neeraj Kumar Singh\footnotemark[1], Pratinav Seth\footnotemark[1]  \\
  \texttt{\{neeraj.singh, pratinav.seth\}@aryaxai.com} \\
  \\
  AryaXAI \\
}

\begin{document}
\maketitle
\begin{abstract}
The rapid growth of AI has led to more complex deep learning models, often operating as opaque "black boxes" with limited transparency in their decision-making. 
This lack of interpretability poses challenges, especially in high-stakes applications where understanding model output is crucial. 
This work highlights the importance of interpretability in fostering trust, accountability, and responsible deployment. 
To address these challenges, we introduce DLBacktrace, a novel, model-agnostic technique designed to provide clear insights into deep learning model decisions across a wide range of domains and architectures, including MLPs, CNNs, and Transformer-based LLM models. 
We present a comprehensive overview of DLBacktrace and benchmark its performance against established interpretability methods such as SHAP, LIME, and GradCAM. 
Our results demonstrate that DLBacktrace effectively enhances understanding of model behavior across diverse tasks.
DLBacktrace is compatible with models developed in both PyTorch and TensorFlow, supporting architectures such as BERT, ResNet, U-Net, and custom DNNs for tabular data. 
The library is open-sourced and available at \url{https://github.com/AryaXAI/DLBacktrace}.
\end{abstract}
\section{Introduction}
Deep learning models, including large language models (LLM) such as ChatGPT \cite{chatgpt}, LLaMA \cite{Touvron2023LLaMAOA}, and Google’s Gemini, have achieved remarkable performance in various applications such as language generation, image recognition, and complex reasoning tasks. 
However, these sophisticated systems remain largely opaque, offering limited insight into the decision-making processes that underpin their predictions.
This opacity poses significant challenges, particularly in high-stakes domains such as healthcare, finance, and law enforcement.
In these areas, understanding model rationales is not merely desirable—it is often a regulatory and ethical imperative.
For example, healthcare professionals must trust and understand AI-assisted diagnostics before implementing treatments, and regulatory frameworks such as the EU GDPR (Article 22) mandate explainability in automated decisions that affect individuals.

The demand for responsible and trustworthy AI has led to the rise of eXplainable AI (XAI) research, aiming to align models with ethical standards, ensure fairness, transparency, and accountability, and increase public trust \cite{Singh2024RethinkingII, Kaur2022SensibleAR, madsen2024interpretability, Tull2024TowardsCI}. Similarly, AI safety research underscores that explainability helps identify and mitigate risks such as reward hacking and catastrophic forgetting, guiding the development of robust systems \cite{Amodei2016ConcretePI, doi:10.1073/pnas.1611835114, Leike2018ScalableAA}.
Various methods have emerged to address interpretability. Model-agnostic approaches like LIME \cite{Ribeiro2016WhySI} and SHAP \cite{Lundberg2017AUA} assign feature importance scores. However, these are often costly, sensitive to data perturbations, and struggle with high-dimensional data such as images and text. Similarly, gradient-based methods like Grad-CAM and Integrated Gradients \cite{Sundararajan2017AxiomaticAF}, can highlight influential regions or tokens, but face architectural constraints, dependence on carefully chosen baselines, and complex interpretation of attention weights. 

Interpretability frameworks often lack unified metrics, complicating comparisons, and sometimes provide conflicting explanations, underscoring the need for standardized evaluation criteria \cite{Krishna2022TheDP, Dinu2020ChallengingCI, Sixt2019WhenEL}. Compliance with regulatory standards amplifies the importance of transparency. In finance, explainable credit scoring, loan approvals, and fraud detection are increasingly mandated \cite{Weber2023ApplicationsOE}. In healthcare, explainable models ensure the trust needed to integrate AI-driven diagnostics responsibly. The complexity of LLMs, in particular, necessitates tailored interpretability strategies to address challenges like hallucinations, computational constraints, and scale, bringing user-centered methods to the forefront \cite{Calderon2024OnBO}.

To address the growing need for effective interpretability methods in deep learning, we present \textbf{DLBacktrace}, a model-agnostic approach that works independently of auxiliary models or baselines. DLBacktrace ensures consistent and reliable explanations across various architectures and data types, including images, text, and tabular data. By tracing relevance from the output back to the input, it assigns relevance scores across layers to highlight feature importance, information flow, and potential biases in predictions. In this work, we make the following contributions:
\begin{itemize}
    \item We introduce DLBacktrace, a robust, model-agnostic method for deep learning interpretability.
    \item We provide a detailed explanation of the approach behind DLBacktrace, addressing challenges related to stability and consistency across different model architectures.
    \item We demonstrate the applicability of DLBacktrace across different domains, including text, image, and tabular data, paving the way for its broader adoption in responsible AI applications.
\end{itemize}
\begin{figure*}
    \centering
    \includegraphics[width=0.72\linewidth]{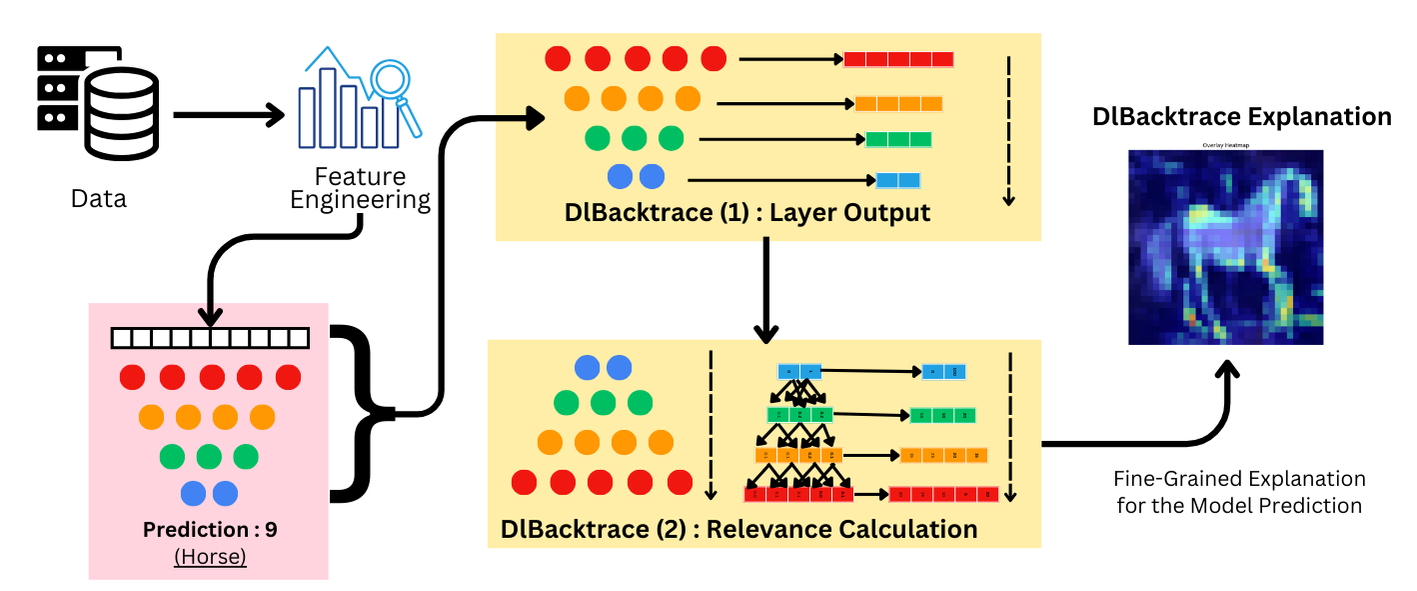}
    \caption{DLBacktrace Workflow: Generating Fine-Grained Explanations from Pre-Trained Models and Test Time Instance.}
    \label{fig:enter-label}
\end{figure*}

\section{Relevant Literature}

\subsection{Importance of eXplainable AI (XAI)}

\subsubsection{XAI for Responsible and Trustworthy AI}

Responsible AI ensures AI systems align with ethical standards, particularly in sectors like healthcare, finance, and law enforcement, where decisions impact individuals and communities. It focuses on fairness, transparency, accountability, privacy, and ethical alignment to minimize bias, protect rights, and build trust.
While progress has been made through regulations like the EU's GDPR and tools such as SHAP, LIME, and Grad-CAM, challenges remain. Current methods provide limited insights, especially with high-dimensional data, and are not always suitable for real-time use. Moreover, rapid AI advancements often outpace regulatory efforts, creating oversight gaps, particularly with large language models (LLMs).
\cite{madsen2024interpretability} advocate for moving beyond traditional interpretability techniques, proposing three approaches that focus on faithful model explanations: models with inherent faithfulness measures, those trained to offer faithful explanations, and self-explaining models.
\cite{Singh2024RethinkingII} address interpretability challenges in LLMs, offering methods tailored to their scale and complexity. 
\cite{Tull2024TowardsCI} introduce a category theory-based framework to unify interpretability approaches. 
\cite{Dinu2020ChallengingCI} critique feature attribution methods, highlighting the need for more reliable evaluations. 
\cite{Kaur2022SensibleAR} apply sensemaking theory to AI, advocating for explanations that align with human cognitive processes to improve trust and accountability.

\subsubsection{XAI for Safe AI}
AI safety ensures that AI systems are predictable, controllable, and aligned with human values, particularly in high-stakes areas like healthcare, autonomous vehicles, and infrastructure. A key aspect of AI safety is explainability, which allows developers, users, and regulators to understand system behavior, detect risks, and implement safeguards. Core principles of AI safety include robustness, reliability, alignment with human intentions, and the use of explainability to clarify decision-making.
Explainability is vital for risk mitigation in dynamic environments, where understanding AI behavior enables safe interventions. For instance, \cite{Amodei2016ConcretePI} highlights challenges like reward hacking and exploration hazards in reinforcement learning, where explainability techniques help analyze reward structures and behavior patterns to prevent unintended actions. 
Catastrophic forgetting, where models forget prior knowledge when learning new tasks, presents risks in sequential learning \cite{doi:10.1073/pnas.1611835114}. XAI methods can identify vulnerable areas in models, supporting mechanisms that preserve critical information. 
Explainability also aids in improving adversarial robustness by revealing vulnerabilities, guiding defenses that enhance safety.

\subsubsection{XAI for Regulatory AI}
Explainable AI (XAI) is essential for regulatory compliance, promoting transparency, fairness, and accountability in AI-driven decisions across sectors like finance, healthcare, and law. Increasingly, regulatory frameworks require interpretable models to ensure oversight, protect user rights, and uphold ethical standards. Key components of XAI in regulatory contexts include transparent decision-making, model auditability, and bias mitigation. In finance, XAI helps clarify decisions on credit scoring, loan approvals, and fraud detection, reducing regulatory risks and building public trust. \cite{Weber2023ApplicationsOE} reviews XAI applications in finance, emphasizing transparency's role in regulatory compliance. 
As large language models (LLMs) gain prominence, interpretability in NLP becomes crucial. \cite{Calderon2024OnBO} examines how interpretability methods align with stakeholder needs, highlighting the importance of stakeholder-centered frameworks for responsible AI deployment.
In healthcare, XAI supports patient safety and ethical standards by helping providers interpret AI-driven diagnoses and treatments.
\cite{Singh2024RethinkingII} explores how LLMs can provide interactive, natural language explanations to improve understanding of complex AI behaviors, addressing challenges like hallucinations and computational costs.
These studies underscore XAI's role in enhancing regulatory compliance by fostering transparency, accountability, and fairness, ensuring ethical AI use and facilitating oversight across regulated sectors.

\subsection{Explainability Methods}
\subsubsection{Tabular Data}
In high-stakes fields like finance and healthcare, machine learning models such as regression and decision trees are often preferred over deep learning models, which are seen as "black boxes." Explainable algorithms like LIME~\cite{Ribeiro2016WhySI} and SHAP~\cite{Lundberg2017AUA} address this by enhancing interpretability. LIME creates simple models around specific data points to highlight key features, while SHAP assigns importance scores to features, providing both global and local explanations.
However, both methods have limitations. LIME’s random sampling can lead to inconsistent explanations, and SHAP can be computationally expensive for large datasets. Additionally, SHAP's model-agnostic approach may be less effective for complex Deep Neural Networks.
\subsubsection{Image Data}
For image data, gradient-based methods such as GradCAM~\cite{Selvaraju2016GradCAMVE}, Vanilla Gradient~\cite{Simonyan2013DeepIC}, SmoothGrad~\cite{Smilkov2017SmoothGradRN}, and Integrated Gradients~\cite{Sundararajan2017AxiomaticAF} are widely used to interpret model predictions. GradCAM generates heatmaps based on gradients, while Vanilla Gradient faces the "saturation problem." SmoothGrad improves clarity by adding noise to gradients, and Integrated Gradients address saturation but require more computation.
\subsubsection{Textual Data}
For text-based tasks, methods like LIME~\cite{Ribeiro2016WhySI} and SHAP~\cite{Lundberg2017AUA} are common for interpreting text classification models, alongside gradient-based methods like GradCAM~\cite{Selvaraju2016GradCAMVE} and Integrated Gradients~\cite{Sundararajan2017AxiomaticAF}. For text generation, challenges such as tokenization effects and randomness are noted by \cite{Amara2024ChallengesAO}, and techniques like LACOAT~\cite{Yu2024LatentCE} improve interpretability by clustering word representations. \cite{Zhao2023IncorporatingAI} introduces a soft-erasure method to improve faithfulness in NLP explanations, while \cite{Edin2024NormalizedAF} proposes Normalized AOPC to standardize feature attribution metrics.
Mechanistic interpretability is gaining traction in LLMs, focusing on understanding how individual components, such as neurons and attention heads, contribute to decision-making \cite{olah2020zoom,Nanda2023ProgressMF}.
\subsubsection{Metrics for Benchmarking Explainability}
Recent work has focused on developing robust frameworks for evaluating the quality of XAI methods.
Quantus~\cite{Hedstrm2022QuantusAE} provides a modular toolkit for evaluating neural network explanations, while BEExAI~\cite{Sithakoul2024BEExAIBT} focuses on tabular data, offering metrics like local fidelity. Despite these advancements, challenges like standardization, usability, and fairness remain, highlighting the need for continued research in XAI evaluation.
\begin{figure}[pt]
    \centering
    \begin{subfigure}{0.39\columnwidth}
        \centering
        \includegraphics[height=8.1cm]{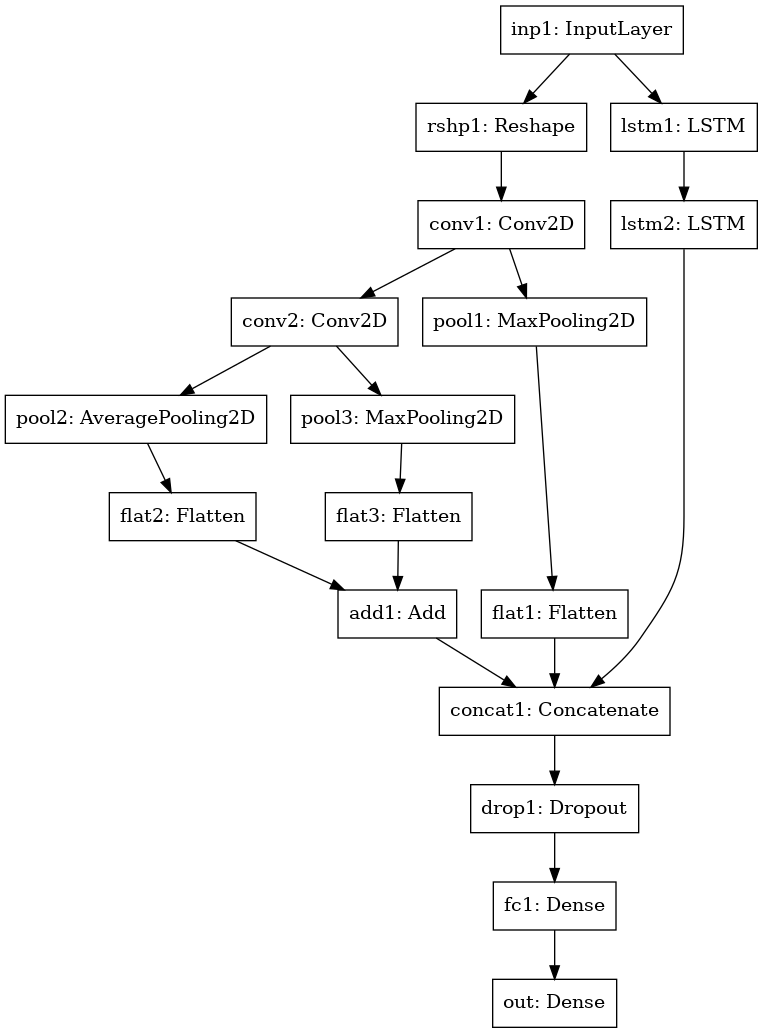}
        \caption{Sample Network}
        \label{fig:image1}
    \end{subfigure}
    \begin{subfigure}{0.59\columnwidth}
        \centering
        \includegraphics[height=8.1cm]{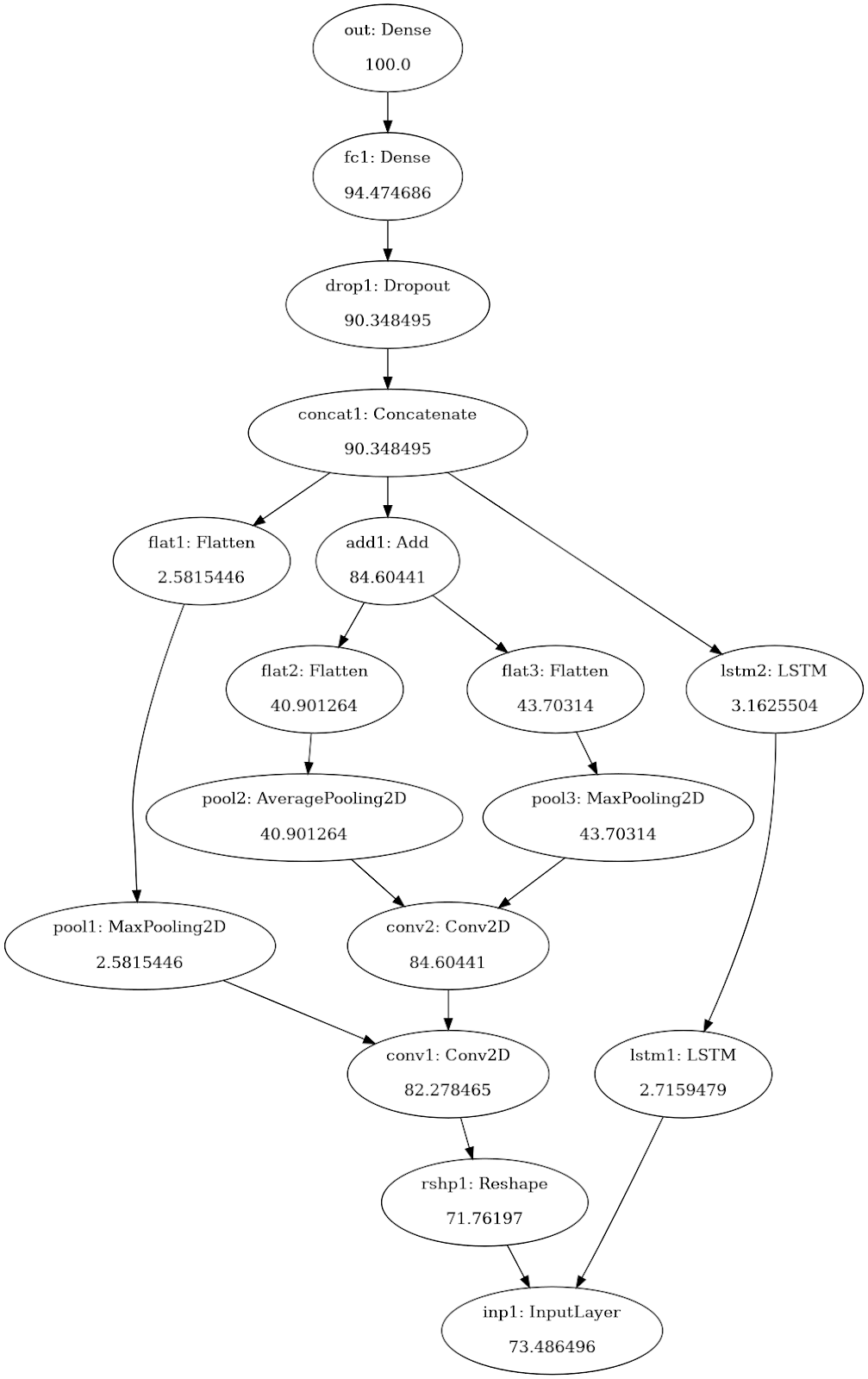}
        \caption{Relevance for Sample Network}
        \label{fig:image2}
    \end{subfigure}
    \caption{Illustration Depicting DLBacktrace Calculation for a Sample Network}
    \label{fig:two_images}
\end{figure}
\section{DLBacktrace}
\subsection{Introduction}
\textbf{DLBacktrace} is a technique for analyzing neural networks that involves tracing the relevance of each component from the output back to the input revealing the feature importance of input, the importance of each layer, information flow, and biases. It operates independently of external algorithms or datasets, ensuring consistency and deterministic results. Relevance is computed directly from the network, making DLBacktrace ideal for live environments and training workflows, with improved interpretation and validation. 
Utilizing DLBacktrace enables the extraction of local relevance for each sample, highlighting the importance unique to that specific instance. By aggregating the normalized scores across multiple samples, it effectively captures the global relevance. 
The procedure for relevance calculation for a given input sample and model (with trained weights) as illustrated in Figure ~\ref{fig:two_images} is as follows:
\begin{enumerate}
    \item Construct a graph from the model weights and architecture with output nodes as root and input nodes as leaves.
    \item Propagate relevance in a breadth-first manner, starting at the root.
    \item The propagation completes when all leaves (input nodes) have been assigned relevance.
\end{enumerate}
\subsection{DLBacktrace Modes}
\textbf{DLBacktrace} works in two modes: Default and Contrastive. 
In \textbf{Default Mode}, a single relevance score is distributed proportionally across positive and negative components, influenced by saturation effects in activation functions.
\textbf{Contrastive Mode} assigns dual relevance scores (positive and negative) to each unit, separating supportive and detractive contributions. This mode facilitates deeper analysis of negatively impactful features and enables bias detection or counterfactual evaluation in applications like sentiment analysis. 

\subsection{Algorithm}

Context: Every deep learning model consists of multiple layers. Each layer has a variation of the basic operation: 

\[
y = \Phi(Wx + b)
\]

where, $\Phi$ = activation function , $W$ = weight matrix of the layer , $b$ = bias , $x$ = input , $y$ = output .

This can be further organized as:
\[
y = \Phi(X_p + X_n + b)
\]
where, $X_p = \sum W_i x_i \quad \forall  W_i x_i > 0$ and $X_n = \sum W_i x_i \quad \forall  W_i x_i < 0$

\subsubsection{Default Mode : }

If the relevance associated with $y$ are $r_{yp}$ $\And$ $r_{yn}$ and with $x$ are $r_{xp}$ $\And$ $r_{xn}$, then for the $j$th unit in $y$, we compute:
\begin{equation}
T_j = X_{pj} + X_{nj} + b_j
\end{equation}
\begin{equation}
R_{pj} = \frac{X_{pj}}{T_j} r_{yj}, \quad R_{nj} = \frac{X_{nj}}{T_j} r_{yj}, \quad R_{bj} = \frac{b_j}{T_j} r_{yj}
\end{equation}
$R_{pj}$ and $R_{nj}$ are distributed among $x$ as follows:
\begin{equation}
\footnotesize
r_{xij} = 
\begin{cases} 
    \frac{W_{ij} x_{ij}}{X_{pj}} R_{pj} & \text{if } W_{ij} x_{ij} > 0 \\
    0 & \text{if } W_{ij} x_{ij} > 0 \text{ and } \Phi \text{ is saturated on negative end} \\
    \frac{-W_{ij} x_{ij}}{X_{nj}} R_{nj} & \text{if } W_{ij} x_{ij} < 0 \\
    0 & \text{if } W_{ij} x_{ij} < 0 \text{ and } \Phi \text{ is saturated on positive end} \\
    0 & \text{if } W_{ij} x_{ij} = 0
\end{cases}
\end{equation}
The total relevance at layer $x$ is:
\begin{equation}
r_{x} = \sum_i r_{xi}
\end{equation}
\subsubsection{Contrastive Mode : }
If the relevance associated with $y$ are $r_{yp}$ $\And$ $r_{yn}$ and with $x$ are $r_{xp}$ $\And$ $r_{xn}$, then for the $j$th unit in $y$, we compute:
\begin{equation}
T_j = X_{pj} + X_{nj} + b_j
\end{equation}
We then determine $R_{pj}$, $R_{nj}$, and Relevance Polarity as described in 
Algorithm~\ref{alg:cont1}.
\begin{algorithm}[H]
\footnotesize
\caption{Determine $R_{pj}$, $R_{nj}$, and relevance polarity in Contrastive Mode}
\begin{algorithmic}[1]
\If{$T_j > 0$}
    \If{$r_{ypj} > r_{ynj}$}
        \State $R_{pj} \gets r_{ypj}$
        \State $R_{nj} \gets r_{ynj}$
        \State $\text{relevance\_polarity} \gets 1$
    \Else
        \State $R_{pj} \gets r_{ynj}$
        \State $R_{nj} \gets r_{ypj}$
        \State $\text{relevance\_polarity} \gets -1$
    \EndIf
\Else
    \If{$r_{ypj} > r_{ynj}$}
        \State $R_{pj} \gets r_{ynj}$
        \State $R_{nj} \gets r_{ypj}$
        \State $\text{relevance\_polarity} \gets -1$
    \Else
        \State $R_{pj} \gets r_{ypj}$
        \State $R_{nj} \gets r_{ynj}$
        \State $\text{relevance\_polarity} \gets 1$
    \EndIf
\EndIf
\end{algorithmic}
\label{alg:cont1}
\end{algorithm}
Afterwards, $R_{pj}$ and $R_{nj}$ are distributed among $x$ as described in Algorithm~\ref{alg:cont2}.
\begin{algorithm}[H]
\footnotesize
\caption{Computation of $r_{xp,ij}$ and $r_{xn,ij}$ based on relevance polarity}
\begin{algorithmic}[1]
\If{$\text{relevance\_polarity} > 0$}
    \State $r_{xp,ij} \gets \frac{W_{ij} x_{ij}}{X_{pj}} R_{pj} \quad \forall W_{ij} x_{ij} > 0$
    \State $r_{xn,ij} \gets \frac{-W_{ij} x_{ij}}{X_{nj}} R_{nj} \quad \forall W_{ij} x_{ij} < 0$
\Else
    \State $r_{xp,ij} \gets \frac{-W_{ij} x_{ij}}{X_{nj}} R_{nj} \quad \forall W_{ij} x_{ij} < 0$
    \State $r_{xn,ij} \gets \frac{W_{ij} x_{ij}}{X_{pj}} R_{pj} \quad \forall W_{ij} x_{ij} > 0$
\EndIf
\end{algorithmic}
\label{alg:cont2}
\end{algorithm}
The total positive and negative relevance at layer $x$ are:
\begin{equation}
r_{xp} = \sum_i r_{xp,i}, \quad r_{xn} = \sum_i r_{xn,i}
\end{equation}
\subsection{Relevance for Attention Layers:}
Most modern AI models rely on the attention mechanism \cite{Vaswani2017AttentionIA}, which has been extended in our work to support the DLBacktrace algorithm.The attention mechanism allows the model to focus on relevant parts of the input sequence by dynamically adjusting the importance of elements for predictions. The attention function is defined as:
\begin{equation}
    Attention(Q, K, V) = \text{softmax}\left(\frac{QK^T}{\sqrt{d_k}}\right)V
\end{equation}
For Multi-Head Attention, the function is:
\begin{equation}
    MultiHead(Q, K, V) = \textbf{Concat}\left(head_1, \ldots, head_n\right)W^O
\end{equation}
\begin{equation}
head_i = \text{Attention}\left(QW_i^Q, KW_i^K, VW_i^V\right)
\end{equation}
Where, \( Q, K, V \): Query, Key, and Value matrices, \( W_i^Q, W_i^K, W_i^V \): Weight matrices for each head,\( W^O \): Weight matrix for combining all heads after concatenation and \text{Concat}: Concatenates the outputs from all heads .

\subsubsection{Algorithm}

Suppose the input to the attention layer is \( x \) and the output is \( y \). The relevance associated with \( y \) is \( r_y \). To compute the relevance using the DLBacktrace, we use the steps as indicated in Algorithm~\ref{alg:cont3} below:

\begin{algorithm}[H]
    \caption{Relevance Propagation for Attention Layers}
    \begin{algorithmic}[1]
        \State \textbf{Input:} \( x \) (input to attention layer)
        \State \textbf{Output:} \( r_y \) (relevance associated with \( y \))
        
        \State We calculate the relevance \( r_O \) of \( \textbf{Concat}\left(head_1, head_2, \ldots, head_n\right) \).\textit{Where \( r_O \) represents the relevance from the linear projection layer of the Attention module.}

        \State To compute the relevance of \( QK^T \) and \( V \), use the following formulas:
        \begin{equation}
            r_{QK} = \left(r_O \cdot x_V\right) \cdot x_{QK}
        \end{equation}
        \begin{equation}
            r_V = \left(x_{QK} \cdot r_O\right) \cdot x_{V}
        \end{equation}
        \textit{Here, \( x_{QK} \) and \( x_V \) are the outputs of \( QK^T \) and \( V \), respectively.}

        \State Now that we have \( r_{QK} \), compute the relevance of \( r_Q \) and \( r_K \) as:
        \begin{equation}
            r_Q = \left(r_{QK} \cdot x_Q\right) \cdot x_K
        \end{equation}
        \begin{equation}
            r_K = \left(x_K \cdot r_{QK}\right) \cdot x_Q
        \end{equation}
        \textit{Here, \( x_Q \) and \( x_K \) are the outputs of \( Q \) and \( K \), respectively.}

        \State To compute \( r_{Attn} \), sum up \( r_Q \), \( r_K \), and \( r_V \):
        \begin{equation}
            r_{Attn} = r_Q + r_K + r_V
        \end{equation}
    \end{algorithmic}
\label{alg:cont3}
\end{algorithm}

Note: Any loss of relevance during propagation is due to network bias.




\section{Usage of DLBacktrace Library}
DLBacktrace supports both PyTorch and TensorFlow frameworks. It can be easily installed using pip. Overall Workflow is as illustrated in Figure~\ref{fig:pipeline-fig}.
\begin{lstlisting}
pip install dl-backtrace
\end{lstlisting}
For TensorFlow models, we can import and initialize the library as follows : (assume we have a pre-trained deep learning model)
\begin{lstlisting}
from dl_backtrace.tf_backtrace import DLBacktrace as B
backtrace = B(model=model)
\end{lstlisting}
Similarly, for Pytorch-based models :
\begin{lstlisting}
from dl_backtrace.pytorch_backtrace import DLBacktrace as B
backtrace = B(model=model)
\end{lstlisting}

To compute layer-wise outputs and evaluate relevance as illustrated in Figure~\ref{fig:two_images} : 
\begin{lstlisting}
layer_outputs = backtrace.predict(test_data[0])
relevance = backtrace.eval(layer_outputs, mode='default', scaler=1, thresholding=0.5, task="binary-classification")
\end{lstlisting}

The relevance evaluation process in DLBacktrace involves adjusting four key attributes: \texttt{mode}, \texttt{scaler}, \texttt{thresholding}, \texttt{model-type}, and \texttt{task}. 

The \texttt{mode} attribute defines the method for evaluating relevance (e.g., "default," "contrastive"), determining how layer contributions are measured. The \texttt{model-type} is used for specific model types that require special operations during graph creation (e.g., 'BERT'). 

The \texttt{scaler} attribute controls the scaling of relevance scores; it provides initial relevance value at the last layer (e.g., \texttt{scaler=1}). 

The \texttt{thresholding} attribute sets a cutoff for model confidence values, discarding predictions to select a predicted class below a specified threshold (e.g., \texttt{thresholding=0.5} it is mostly used for segmentation task). 

The \texttt{task} attribute specifies the type of task being performed (e.g., "binary-classification", "regression", "binary-segmentation"), influencing how relevance is computed based on the model task.

\section{Benchmarking}
In this section, we present a comparative study to benchmark our proposed DLBacktrace algorithm against various existing explainability methods. The goal of this evaluation is to assess the effectiveness, robustness, and interpretability of DLBacktrace in providing meaningful insights into model predictions across different data modalities, including tabular, image, and text data. By systematically comparing our approach with established methods, we aim to highlight the advantages and potential limitations of DLBacktrace in the context of explainable artificial intelligence (XAI).

\subsection{Setup}
The experimental setup consists of three distinct data modalities: tabular, image, and text. Each modality is associated with specific tasks, datasets, and model architectures tailored to effectively evaluate the explainability methods. 

For the \textbf{tabular data modality}, we focus on a binary classification task utilizing the Lending Club dataset. This dataset is representative of financial applications, containing features that capture various attributes of borrower profiles. We employ a four-layer Multi-Layer Perceptron (MLP) neural network, which is well-suited for learning from structured data and provides a foundation for assessing the performance of explainability techniques.

In the \textbf{image data modality}, we conduct a multi-class classification task using the CIFAR-10 dataset. This benchmark dataset consists of images across 10 different classes, making it ideal for evaluating image classification algorithms. For this experiment, we utilize a fine-tuned ResNet-34 model, known for its deep residual learning capabilities, which enhances the model's ability to learn intricate patterns and features within the images.

The \textbf{text data modality} involves a binary classification task using the SST-2 dataset, which is focused on sentiment analysis. The dataset consists of movie reviews labeled as positive or negative, allowing for a nuanced evaluation of sentiment classification models. We employ a pre-trained BERT model, which leverages transformer-based architectures to capture contextual relationships in text. This approach facilitates the generation of high-quality explanations for the model's predictions, enabling a thorough assessment of explainability methods in the realm of natural language processing.

\subsection{Metrics}
To assess the effectiveness of explanation methods across various modalities, we utilize different metrics tailored to specific use cases. Further details are provided below:
\begin{figure}[pt]
    \centering
    \begin{subfigure}{0.3\columnwidth}
        \includegraphics[width=\linewidth]{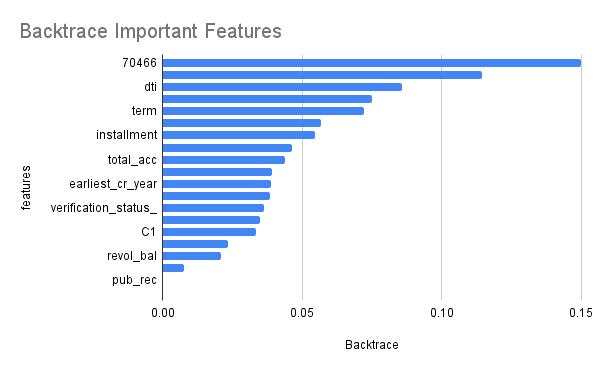}
        \caption{DLBacktrace}
    \end{subfigure}
    \begin{subfigure}{0.3\columnwidth}
        \includegraphics[width=\linewidth]{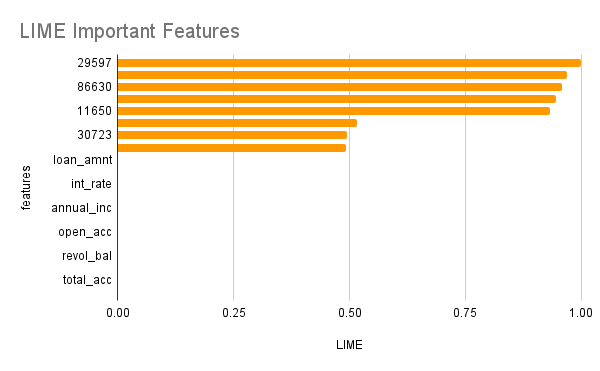}
        \caption{LIME}
    \end{subfigure}
    \begin{subfigure}{0.3\columnwidth}
        \includegraphics[width=\linewidth]{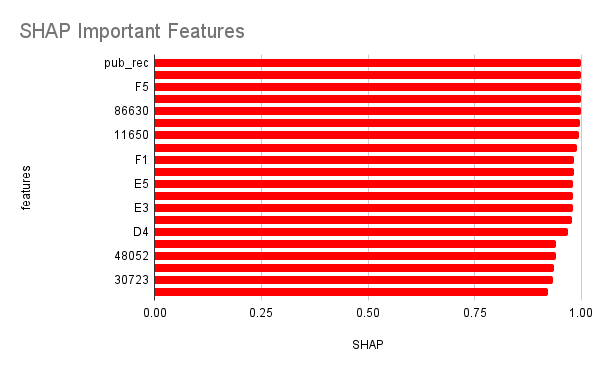}
        \caption{SHAP}
    \end{subfigure}
    \caption{Illustration of Explanations of a Correctly Classified Sample from the Lending Club Dataset where Loan was Fully Paid and was predicted by MLP as Fully Paid.}
    \label{fig:tabill}
\end{figure}
\subsubsection{Tabular Modality:} For Tabular Modality we use MPRT and Complexity.

\textbf{Maximal Perturbation Robustness Test (MPRT):} This metric assesses the extent of perturbation that can be applied to an input before there are significant changes in the model's explanation of its decision. It evaluates the stability and robustness of the model's explanations rather than solely its predictions.

\textbf{Complexity Metric:} This metric quantifies the level of detail in a model's explanation by analyzing the distribution of feature contributions.

\begin{figure}[pt]
    \centering
    \begin{subfigure}{0.162\linewidth}
        \includegraphics[width=\linewidth]{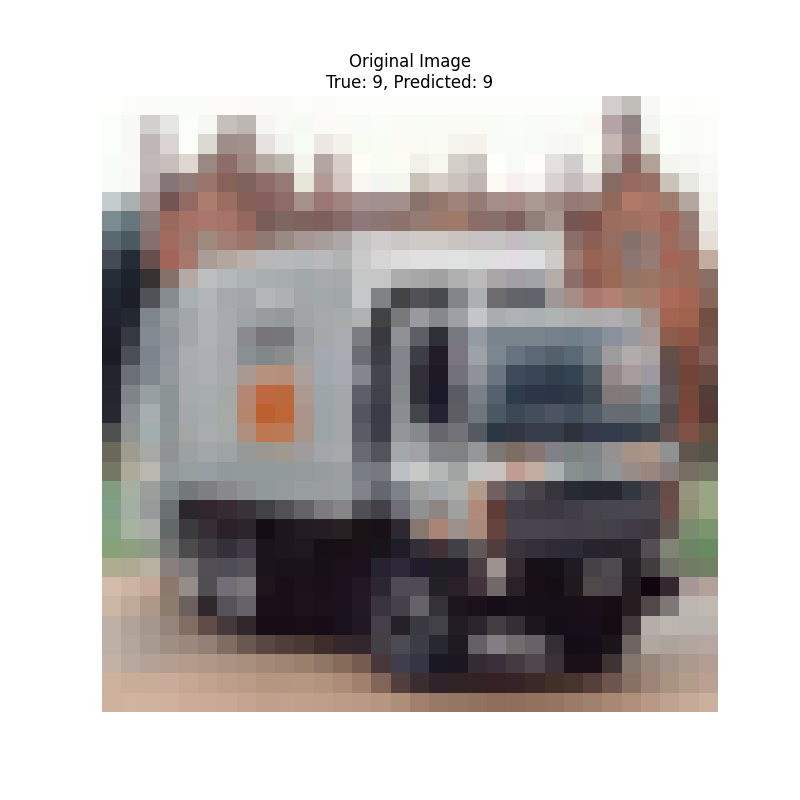}
        \caption{Original Image}
    \end{subfigure}
    \begin{subfigure}{0.162\linewidth}
        \includegraphics[width=\linewidth]{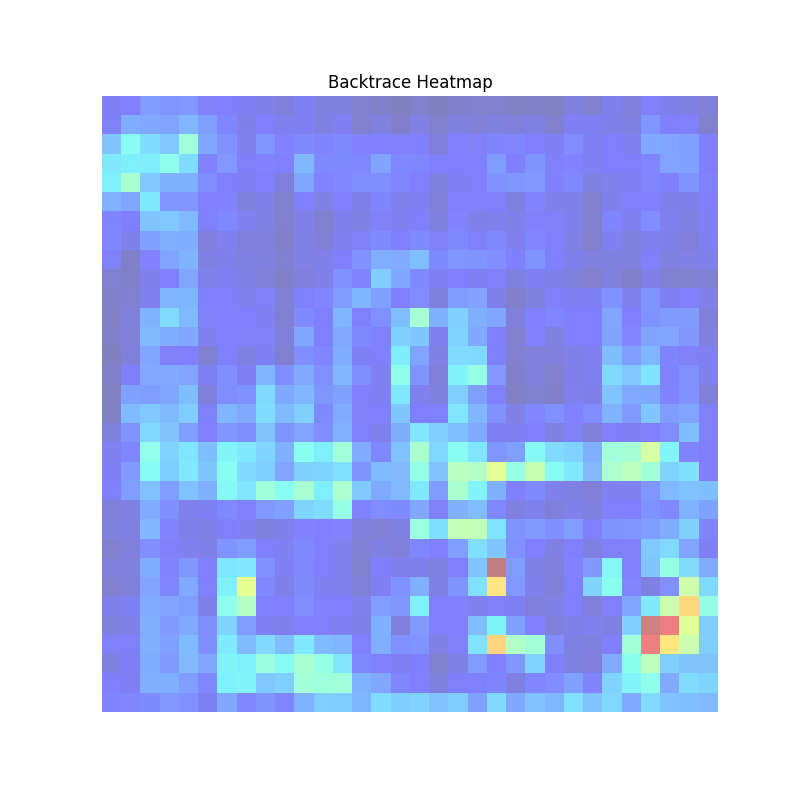}
        \caption{Backtrace}
    \end{subfigure}
    \begin{subfigure}{0.162\linewidth}
        \includegraphics[width=\linewidth]{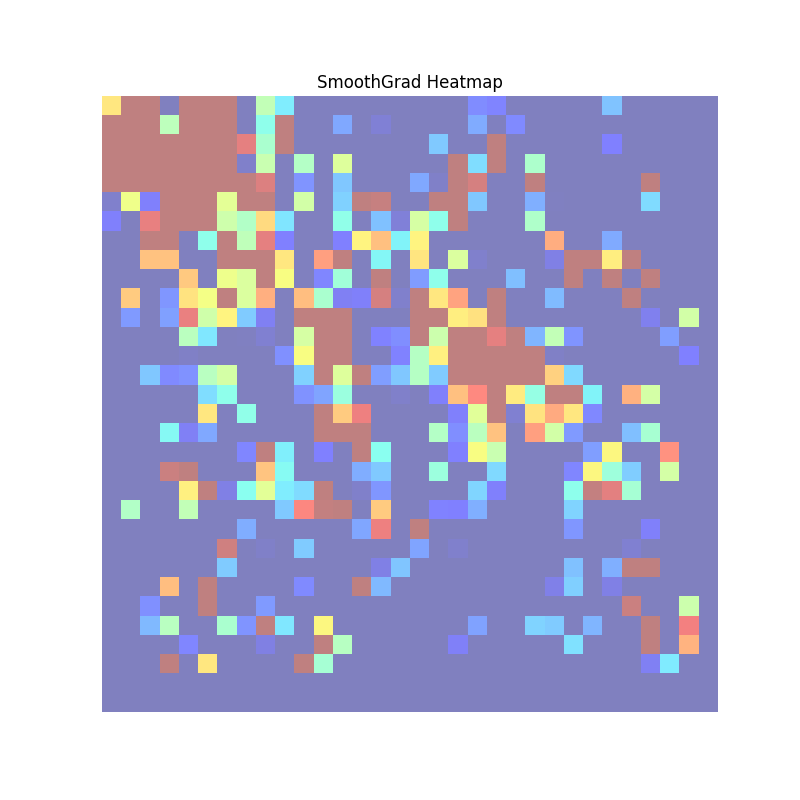}
        \caption{Smooth Grad}
    \end{subfigure}
    \begin{subfigure}{0.162\linewidth}
        \includegraphics[width=\linewidth]{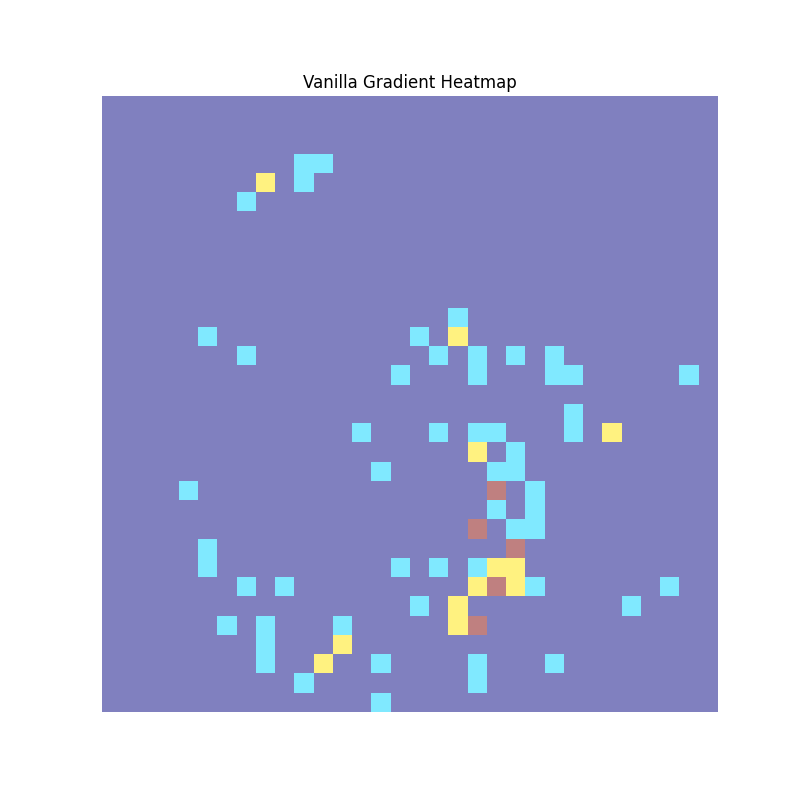}
        \caption{Vanilla Gradient}
    \end{subfigure}
    \begin{subfigure}{0.162\linewidth}
        \includegraphics[width=\linewidth]{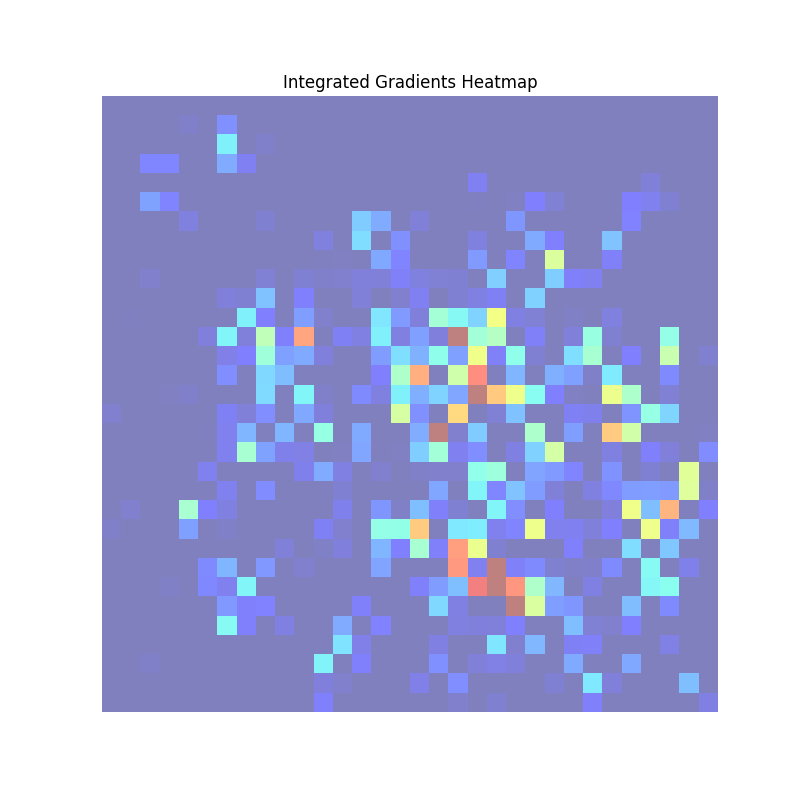}
        \caption{IG}
    \end{subfigure}
    \begin{subfigure}{0.162\linewidth}
        \includegraphics[width=\linewidth]{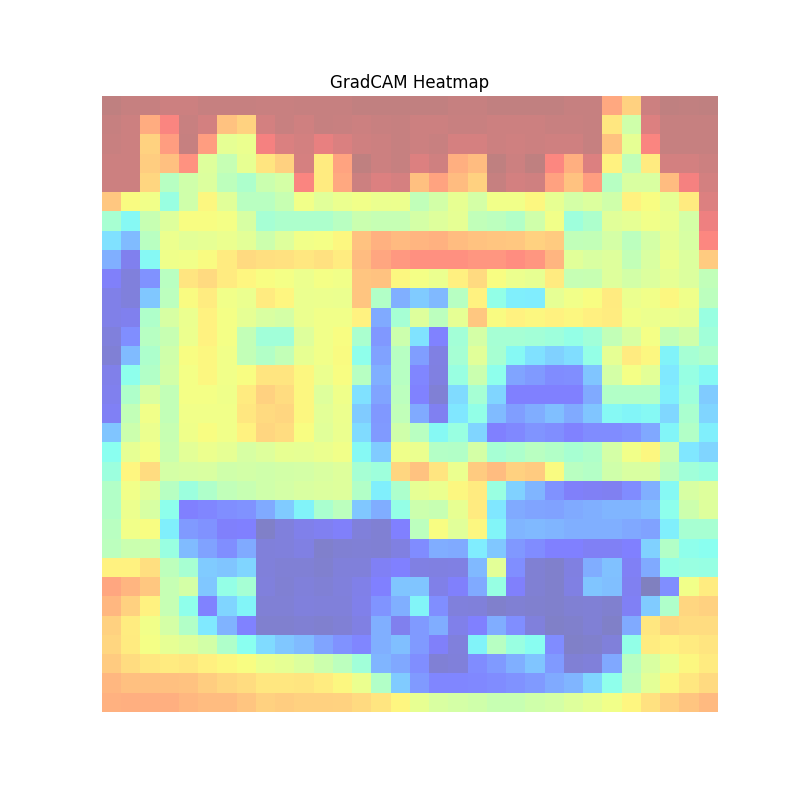}
        \caption{GradCAM}
    \end{subfigure}
    \caption{Visualizing ResNet's decisions on a Truck image of CIFAR10 Dataset using various explanation methods.}
    \label{fig: Multi-class Classification Interpretations-2}
\end{figure}

\begin{figure}[pt]
    \centering
    \includegraphics[width=0.8\columnwidth]{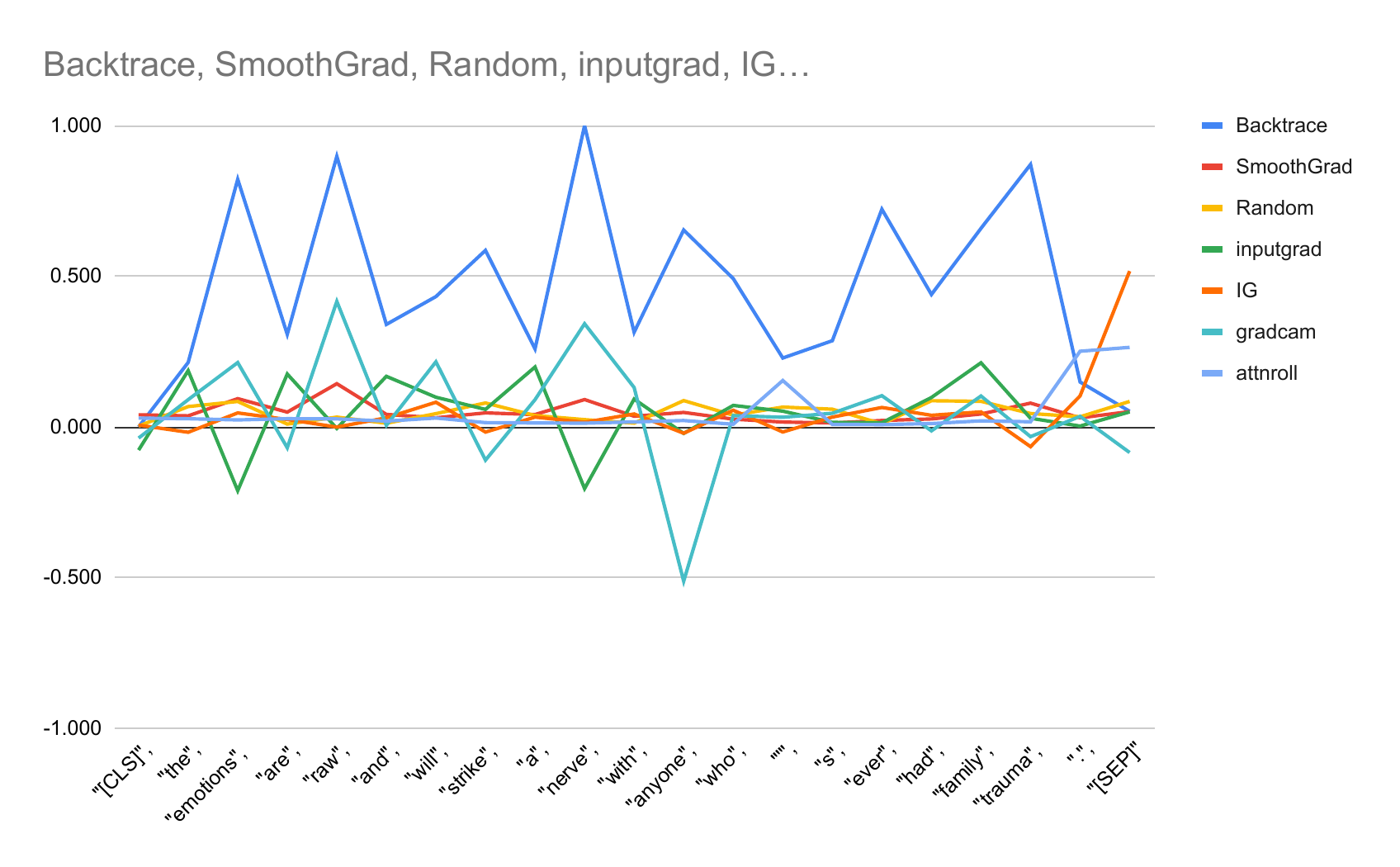}
    \caption{Explanations by different methods for model decision making for Sentiment Analysis for a sample from SST Dataset.  \textbf{Input Text:} The emotions are raw and will strike a nerve with anyone who ever had family trauma. \textbf{Prediction: 1} and \textbf{Label: 1}}
    \label{fig:bert_text}
\end{figure}
\subsubsection{Image Modality} For Image Modality we use Faithfulness Correlation, Max-Sensitivity and Pixel Flipping .

\textbf{Faithfulness Correlation:} Faithfulness Correlation \cite{Bhatt2020EvaluatingAA} metric evaluates the degree to which an explanation aligns with the model’s behavior by calculating the correlation between feature importance and changes in model output resulting from perturbations of key features.

\textbf{Max Sensitivity:} Max-Sensitivity \cite{Yeh2019OnT}, is a robustness metric for explainability methods that evaluates how sensitive explanations are to small perturbations in the input. Using a Monte Carlo sampling-based approximation, it measures the maximum change in the explanation when slight random modifications are applied to the input. Formally, it computes the maximum distance (e.g., using \(L_1\), \(L_2\), or \(L_\infty\) norms) between the original explanation and those derived from perturbed inputs. 

\textbf{Pixel Flipping:} Pixel Flipping  method involves perturbing significant pixels and measuring the degradation in the model’s prediction, thereby testing the robustness of the generated explanation.

\subsubsection{Text Modality}
To evaluate the textual modality, we use the Token Perturbation for Explanation Quality (ToPEQ) metric, which assesses the robustness of model explanations by analyzing the impact of token perturbations. 

\textbf{LeRF AUC (Least Relevant First AUC):} This metric evaluates how gradually perturbing the least important features (tokens) affects the model's confidence. The AUC measures the model’s response as the least relevant features are replaced with a baseline (e.g., [UNK]), indicating the degree to which the model relies on these features.

\textbf{MoRF AUC (Most Relevant First AUC):} This metric measures how quickly the model's performance deteriorates when the most important features are perturbed first. The AUC represents the decrease in the model's confidence as the most relevant tokens are removed, revealing the impact of these key features on the prediction.

\textbf{Delta AUC:} This metric represents the difference between LeRF AUC and MoRF AUC. It reflects the model's sensitivity to the removal of important features (MoRF) compared to less important ones (LeRF). A larger delta suggests that the explanation method effectively distinguishes between important and unimportant features.
\begin{table}[pt]
\centering
\caption{Explaination Performance metrics for explanation methods - LIME,SHAP and DLBacktrace, including Mean values and feature contributions across different layers (fc1 to fc4). Lower values in MPRT and Complexity indicate better performance.}
\begin{tabular}{|ccccccc|}
\hline
\multicolumn{1}{|c|}{\multirow{2}{*}{\textbf{Method}}} & \multicolumn{5}{c|}{\textbf{MPRT (↓)}}                                                                                                                                             & \multirow{2}{*}{\textbf{\begin{tabular}[c]{@{}c@{}}Complexity\\ (↓)\end{tabular}}} \\ \cline{2-6}
\multicolumn{1}{|c|}{}                                 & \multicolumn{1}{c|}{\textbf{Mean}} & \multicolumn{1}{c|}{\textbf{fc1}} & \multicolumn{1}{c|}{\textbf{fc2}} & \multicolumn{1}{c|}{\textbf{fc3}} & \multicolumn{1}{c|}{\textbf{fc4}} &                                                                                    \\ \hline
LIME                                                   & 0.933                              & 0.934                             & 0.933                             & 0.933                             & 0.933                             & 2.57                                                                               \\
SHAP                                                   & 0.684                              & 0.718                             & 0.65                              & 0.699                             & 0.669                             & \textbf{1.234}                                                                     \\
{DLBacktrace}                               & \textbf{0.562}                     & \textbf{0.579}                    & \textbf{0.561}                    & \textbf{0.557}                    & \textbf{0.552}                    & 2.201                                                                              \\ \hline
\end{tabular}
\label{tab:performance_metrics_tabular}
\end{table}
\subsection{Experiments}
\subsubsection{Tabular Modality}
We evaluated 1,024 samples from the test set of the Lending Club dataset, using a fine-tuned MLP checkpoint that attained an accuracy of 0.89 and a weighted average F1 score of 0.87. We assessed DLBacktrace against widely used metrics for tabular data, specifically LIME and SHAP \cite{Lundberg2017AUA}, and employed MPRT for comparison, along with Complexity to examine the simplicity of the model explanations as illustrated in Fig~\ref{fig:tabill}.

The proposed method, \textbf{DLBacktrace}, as demonstrated in Table~\ref{tab:performance_metrics_tabular}, achieves superior performance compared to both LIME and SHAP, evidenced by lower Maximal Perturbation Robustness Test (MPRT) values across various layers. This highlights its improved interpretability and robustness in explainability. However, DLBacktrace exhibits higher computational complexity, reflecting the fine-grained, higher entropy of its explanations. This trade-off suggests the need to balance the quality of interpretability with the simplicity of model explanations.

\begin{table}[pt]
    \centering
    \caption{
Performance metrics of various explanation methods for a subset of CIFAR10 test set samples. Higher values (↑) of Faithfulness Correlation indicate better performance, while lower values (↓) of Max Sensitivity and Pixel Flipping suggest improved robustness. (*) - Indicates the presence of infinite values in some batches, for which a non-infinite mean was used to calculate the final value.}
\begin{tabular}{|cccc|}
\hline
\multicolumn{1}{|c|}{\textbf{\begin{tabular}[c]{@{}c@{}}Explanation\\ Method\end{tabular}}} & \multicolumn{1}{c|}{\textbf{\begin{tabular}[c]{@{}c@{}}Faithfulness \\ Correlation (↑)\end{tabular}}} & \multicolumn{1}{c|}{\textbf{\begin{tabular}[c]{@{}c@{}}Max \\ Sensitivity (↓)\end{tabular}}} & \textbf{\begin{tabular}[c]{@{}c@{}}Pixel\\ Flipping (↓)\end{tabular}} \\ \hline
GradCAM                                                                                     & 0.010                                                                                                 & 1070(*)                                                                              & 0.249                                                                 \\
Vanilla Gradient                                                                            & 0.011                                                                                                 & 154(*)                                                                               & 0.253                                                                 \\
Smooth Grad                                                                                 & 0.018                                                                                                 & 158(*)                                                                                 & 0.252                                                                 \\
Integrated Gradient                                                                         & 0.009                                                                                                 & 169(*)                                                                                 & 0.253                                                                 \\
DLBacktrace                                                                                   & \textbf{0.199}                                                                                        & \textbf{0.617}                                                                               & \textbf{0.199}                                                        \\ \hline
\end{tabular}
    \label{tab:performance_metrics_image}
\end{table}
\begin{table}[pt]
\centering
\caption{Token Perturbation for Explanation Quality metrics for various explanation methods. Lower MoRF AUC values indicate better performance, while higher LeRF AUC and Delta AUC values suggest greater robustness and better differentiation between relevant and irrelevant features.}
\begin{tabular}{|cccc|}
\hline
\multicolumn{1}{|c|}{\textbf{Method}} & \multicolumn{1}{c|}{\textbf{\begin{tabular}[c]{@{}c@{}}MoRF \\ AUC (↓)\end{tabular}}} & \multicolumn{1}{c|}{\textbf{\begin{tabular}[c]{@{}c@{}}LeRF\\ AUC (↑)\end{tabular}}} & \textbf{\begin{tabular}[c]{@{}c@{}}Delta\\ AUC (↑)\end{tabular}} \\ \hline
IG                                    & -6.723                                                                                & 46.756                                                                               & 53.479                                                           \\
Smooth Grad                           & 14.568                                                                                & 38.264                                                                               & 23.696                                                           \\
Attn Roll                             & 16.123                                                                                & 37.937                                                                               & 21.814                                                           \\
DLBacktrace                             & 15.431                                                                                & 30.69                                                                                & 15.259                                                           \\
GradCAM                               & 19.955                                                                                & 21.714                                                                               & 1.759                                                            \\
Random                                & 25.068                                                                                & 25.684                                                                               & 0.616                                                            \\
Input Grad                            & 27.784                                                                                & 19.34                                                                                & -8.444                                                           \\ \hline
\end{tabular}
\label{tab:performance_metrics_text}
\end{table}
\subsubsection{Image Modality}
We conducted an evaluation on 500 samples from the CIFAR-10 test set using a supervised, fine-tuned ResNet-34 model, which achieved a test accuracy of 75.85\%. 
We compared DLBacktrace against several methods as illustrated in Fig.~\ref{fig: Multi-class Classification Interpretations-2}, including Grad-CAM, vanilla gradient, smooth gradient, and integrated gradient. The comparison utilized metrics such as Faithfulness Correlation, Max Sensitivity, and Pixel Flipping.

The Evaluations conducted as shown in Table~\ref{tab:performance_metrics_image} 
reveals that DLBacktrace significantly outperforms traditional methods like Grad-CAM, Vanilla Gradient, Smooth Gradient, and Integrated Gradient across all key metrics. DLBacktrace achieves a superior Faithfulness Correlation score of (0.199), indicating a stronger alignment between its explanations and the model’s behavior. Additionally, it demonstrates robust performance with much lower Max Sensitivity (0.617) and Pixel Flipping (0.199) scores, highlighting its stability against input perturbations and better robustness in preserving the model’s predictive integrity under pixel modifications. Overall, DLBacktrace establishes itself as a more reliable and robust explainability technique for image modality tasks.
\subsubsection{Text Modality}
For the text modality, evaluation was performed on the evaluation set of the SST-2 dataset using a fine-tuned BERT model, achieving an F1 score of 0.926. Since explainable AI (XAI) for BERT and other transformer-based models is relatively new, we employed metrics based on token perturbation for explanation quality, specifically LeRF, MoRF, and Delta AUC, as introduced in \cite{Achtibat2024AttnLRPAL}. We used methods like IG, SmoothGrad, AttnRoll, GradCAM and Input Grad as illustrated in Fig.~\ref{fig:bert_text}.

As shown in Table~\ref{tab:performance_metrics_text}, Integrated Gradients (IG) delivered the strongest performance, achieving the lowest MoRF AUC and the highest LeRF and Delta AUC, underscoring the robustness of its explanations and precise feature attribution. Smooth Grad and Attn Roll also demonstrated commendable performance. DLBacktrace exhibited balanced results across LeRF and Delta AUC metrics, with a MoRF AUC of 15.431, showcasing its ability to provide meaningful explanations while allowing scope for further enhancement. These findings highlight the effectiveness of IG in explainability for transformer-based models while recognizing DLBacktrace's potential as a promising and competitive approach.
\subsubsection{Observations}
The quality of explanations is influenced by both the input data and the model weights. The impact of model performance is significant; low model performance tends to result in unstable explanations characterized by high entropy, while good model performance is associated with stable explanations that are more sparse. Additionally, the inference time for a sample is proportional to both the size of the model and the computational infrastructure used.

\section{Discussion}
DLBacktrace framework offers key advantages in model interpretation. It improves network analysis by addressing the limitations of traditional methods like distribution graphs and heatmaps, which only show node activations without reflecting their impact on the final prediction. It also distinguishes between input data and internal network biases. For feature importance, DLBacktrace assigns weights to input sources, making it easier to understand their influence on predictions, and avoids the issues seen with methods like Integrated Gradients and Shapley values. Additionally, it incorporates uncertainty by evaluating node weight distributions relative to past predictions.

DLBacktrace is versatile and applicable in several areas. It provides local and global feature importance, with global importance averaged across samples. In network analysis, it offers insights into layer relevance, including bias-to-input ratios and activation saturation, guiding network adjustments. The framework also supports fairness and bias analysis by evaluating the influence of sensitive features like gender or age. It helps assess process compliance by comparing feature rankings with business rules or regulations.

\subsection{DLBacktrace Advantages}
\label{sec:dlbacktrace_advantages_appendix}
DLBacktrace has the following advantages over other available tools:
\begin{itemize}
    \item \textbf{No dependence on a sample selection algorithm :} \\The relevance is calculated using just the sample in focus. This avoids deviations in importance due to varying trends in sample datasets.
    \item \textbf{No dependence on a secondary white-box algorithm :} \\The relevance is calculated directly from the network itself. This prevents any variation in importance due to type, hyperparameters, and assumptions of secondary algorithms.
    \item \textbf{Deterministic in nature}\\The relevance scores won’t change on repeated calculations on the same sample. Hence, it can be used in live environments or training workflows as a result of its independence from external factors.
\end{itemize}
\subsubsection{Network Analysis}
\begin{itemize}
    \item Existing solutions involve distribution graphs and heatmaps for any network node based on node activation. 
    \item These are accurate for that specific node but don't represent the impact of that node on the final prediction.
    \item Existing solutions are also unable to differentiate between the impact of input sources versus the internal network biases.
\end{itemize}

\subsubsection{Feature Importance}
\begin{itemize}
    \item With each input source being assigned a fraction of the overall weightage, we can now quantify the dependence of the final prediction on each input source.
    \item We can also evaluate the dependence within the input source as the weight assignment happens on a per-unit basis.
    \item Integrated Gradients and Shapley values are other methods available for calculating feature importance from Deep Learning Models. Both come with caveats and give approximate values:
    \begin{itemize}
        \item Integrated Gradients depends on a baseline sample which needs to be constructed for the dataset and altered as the dataset shifts. This is extremely difficult for high-dimensional datasets.
        \item Shapley Values are calculated on a sample set selected from the complete dataset. This makes those values highly dependent on the selection of data.
    \end{itemize}
\end{itemize}

\subsubsection{Uncertainty}
\begin{itemize}
    \item Instead of just relying on the final prediction score for decision-making, the validity of the decision can now be determined based on the weight distribution of any particular node with respect to the prior distribution of correct and incorrect predictions.
\end{itemize}

\subsection{Applicability}
The DLBacktrace framework is applicable in the following use cases:

\subsubsection{Interpreting the model outcomes using the local and global importance of each feature}
The local importance is directly inferred from the relevance associated with input data layers. For inferring global importance, the local importance of each sample is normalized with respect to the model outcome of that sample. The normalized local importance from all samples is then averaged to provide global importance. The averaging can be further graded based on the various outcomes and binning of the model outcome.

\subsubsection{Network analysis based on the relevance attributed to each layer in the network}
The two modes together provide a lot of information for each layer, such as:
\begin{itemize}
    \item Bias to input ratio
    \item Activation Saturation
    \item Positive and negative relevance (unit-wise and complete layer)
\end{itemize}
Using this information, layers can be modified to increase or decrease variability and reduce network bias. Major changes to the network architecture via complete shutdown of nodes or pathways are also possible based on the total contribution of that component.

\subsubsection{Fairness and bias analysis using the feature-wise importance}
This is in continuation of the global importance of features. Based on the global importance of sensitive features (e.g., gender, age, etc.) and their alignment with the data, it can be inferred whether the model or data has an undue bias towards any feature value.

\subsubsection{Process Compliance based on the ranking of features on local and global levels}
Using the local and global importance of features and ranking them accordingly, it can be determined whether the model is considering the features in the same manner as in the business process it is emulating. This also helps in evaluating the solution’s alignment with various business and regulatory requirements.

\subsubsection{Validating the model outcome}
Every model is analyzed based on certain performance metrics which are calculated over a compiled validation dataset. This doesn’t represent the live deployment scenario. 

During deployment, validation of outcomes is extremely important for complete autonomous systems. The layer-wise relevance can be used to accomplish this. The relevance for each layer is mapped in the vector space of the same dimension as the layer outcome, yet it is linearly related to the model outcome. 

Since the information changes as it passes through the network, the relevance from lower layers, even input layers, can be used to get different outcomes. These outcomes can be used to validate the model outcome. The layers are generally multi-dimensional, for which proximity-based methods or white-box regression algorithms can derive outcomes.

\section{Conclusion}
In this paper, we introduced the \textbf{DLBacktrace}, a new method that significantly improves model interpretability for deep learning. DLBacktrace traces relevance from output back to input, giving clear and consistent insights into which features are important and how information flows through the model. Unlike existing methods, which often rely on changing inputs or using other algorithms, DLBacktrace is stable and reliable, making it especially useful in fields that need high transparency, like finance, healthcare, and regulatory compliance. Our benchmarking results demonstrate that DLBacktrace performs better in terms of robustness and accuracy across various model types, proving it can provide practical insights. Overall, DLBacktrace contributes to the growing field of explainable AI by enhancing model transparency and trustworthiness, promoting responsible AI deployment in critical applications.

\section{Future Works}
Future research on DLBacktrace will focus on expanding its applicability and improving its relevance scoring. Key areas include adapting it for complex models like advanced transformers and multimodal systems, ensuring its effectiveness across different AI applications. Efforts will also be made to reduce inference time for real-time use in production environments such as autonomous systems.
We will explore it's usage for model optimization, including pruning by identifying non-essential components. Additionally, we will also explore it's usage for out-of-distribution (OOD) detection to enhance model robustness.

\section{Ethical Concerns}
DLBacktrace has the potential to support compliance with regulatory requirements, such as the EU GDPR Article 22, which grants individuals the right to an explanation regarding automated decision-making processes that significantly affect them. DLBacktrace helps ensure that AI systems remain accountable and meet these legal standards by providing transparent insights into model decisions. To foster responsible use, it is essential that DLBacktrace is implemented with the appropriate safeguards, including clear communication with users about how their data is being utilized and efforts to prevent any misuse. This study exclusively uses publicly available datasets and tools from open domains, adhering to their respective licensing and usage terms. No personal or sensitive data is involved in the study. Furthermore, the codebase used for this research is already publicly accessible, promoting transparency and reproducibility.
\appendix
\section{Illustrations}
\label{sec:illustrations}
\subsection{Illustrations for Various Tasks}
\subsubsection{Tabular Modality : Binary Classification}
\label{sec:tab-samples}
\begin{figure}[H]
    \centering
    \begin{subfigure}{0.32\linewidth}
        \includegraphics[width=\linewidth]{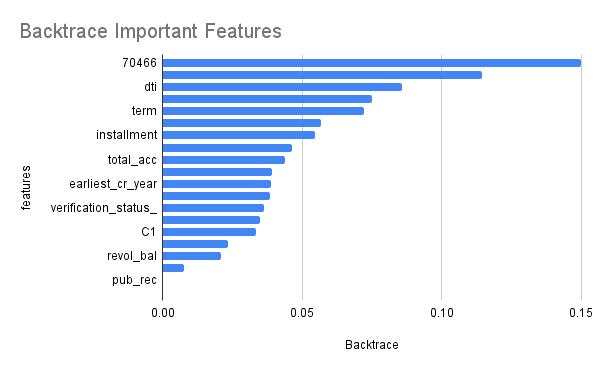}
        \caption{Backtrace}
    \end{subfigure}
    \begin{subfigure}{0.32\linewidth}
        \includegraphics[width=\linewidth]{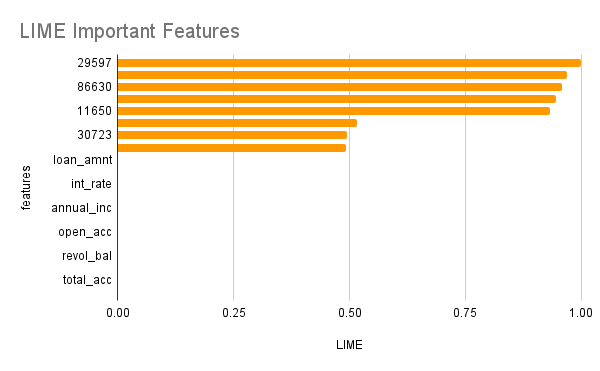}
        \caption{LIME}
    \end{subfigure}
    \begin{subfigure}{0.32\linewidth}
        \includegraphics[width=\linewidth]{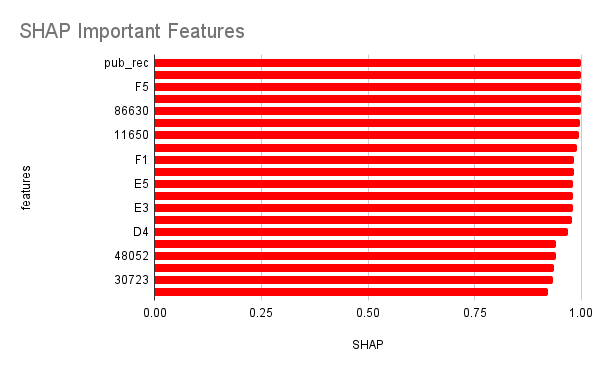}
        \caption{SHAP}
    \end{subfigure}
    \caption{Illustration of Explanations of a Correctly Classified Sample from the Lending Club Dataset where Loan was Fully Paid and was predicted by MLP as Fully Paid.}
    \label{fig: Object Detection Interpretations}
\end{figure}

\begin{figure}[H]
    \centering
    \begin{subfigure}{0.32\linewidth}
        \includegraphics[width=\linewidth]{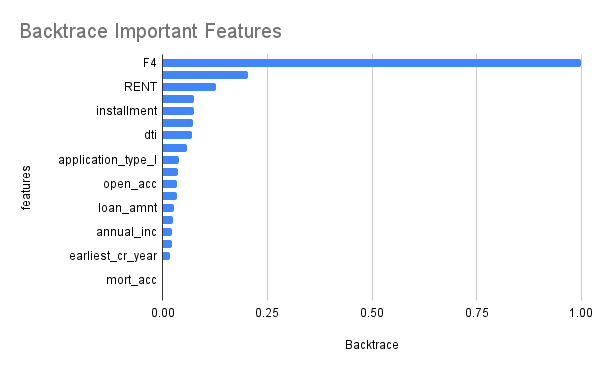}
        \caption{Backtrace}
    \end{subfigure}
    \begin{subfigure}{0.32\linewidth}
        \includegraphics[width=\linewidth]{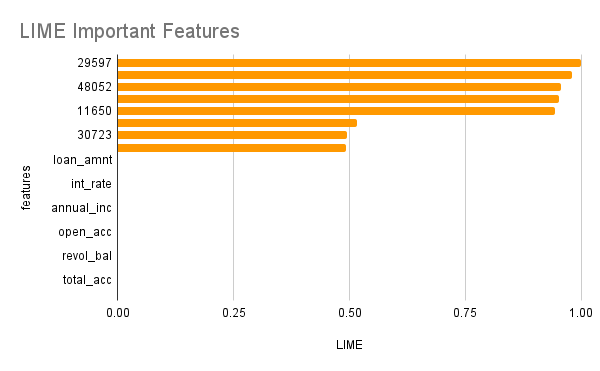}
        \caption{LIME}
    \end{subfigure}
    \begin{subfigure}{0.32\linewidth}
        \includegraphics[width=\linewidth]{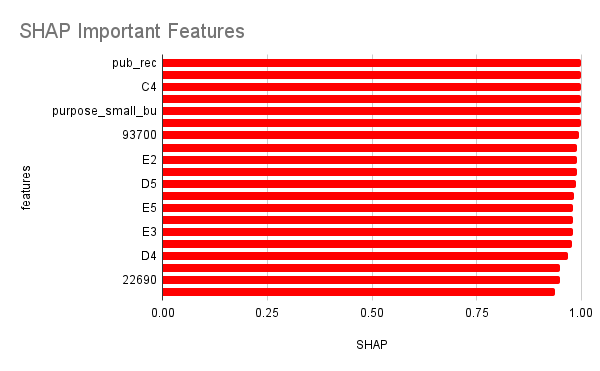}
        \caption{SHAP}
    \end{subfigure}
    \caption{Illustration of Explanations of a Incorrectly Classified Sample from the Lending Club Dataset where Loan was Fully Paid and was predicted by MLP as Charged Off.}
    \label{fig:Fraud_Detection_Interpretations}
\end{figure}

\subsubsection{Image Modality : Multi-Class Classification}
\label{sec:image-samples}
\begin{figure}[H]
    \centering
    \begin{subfigure}{0.162\linewidth}
        \includegraphics[width=\linewidth]{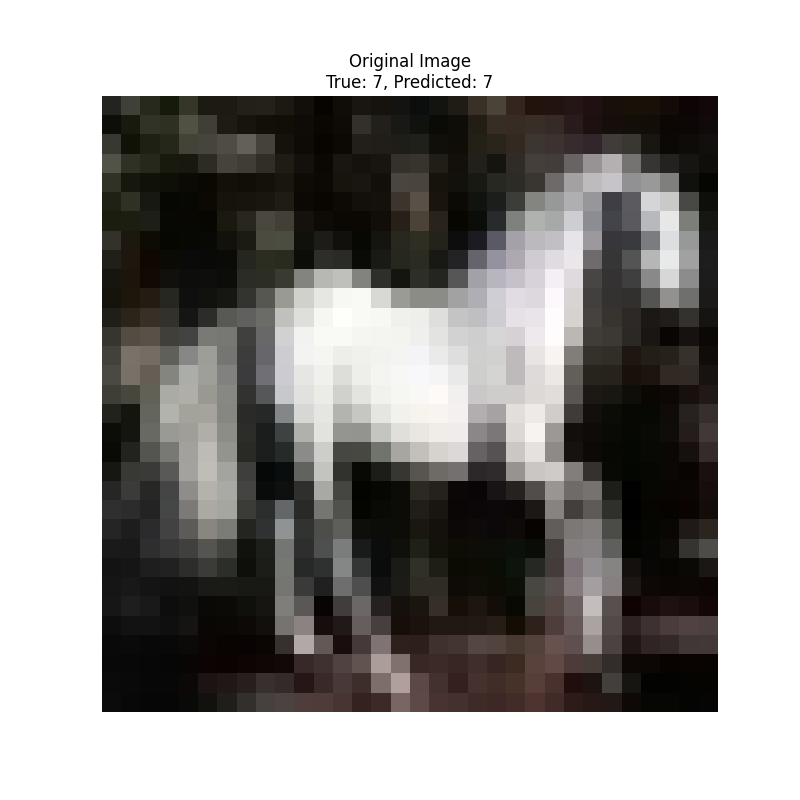}
        \caption{Original Image}
    \end{subfigure}
    \begin{subfigure}{0.162\linewidth}
        \includegraphics[width=\linewidth]{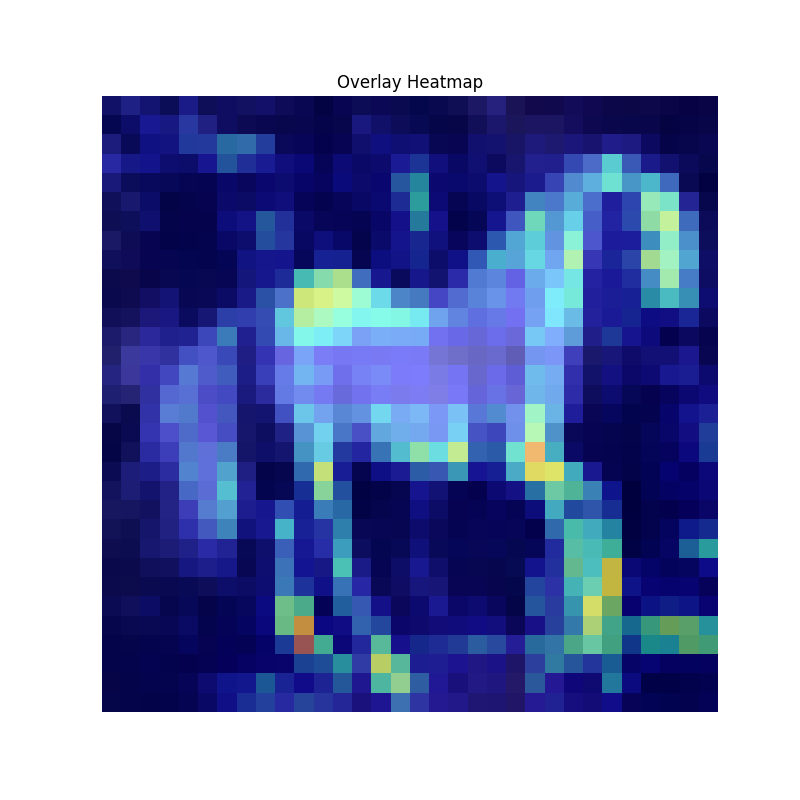}
        \caption{Backtrace}
    \end{subfigure}
    \begin{subfigure}{0.162\linewidth}
        \includegraphics[width=\linewidth]{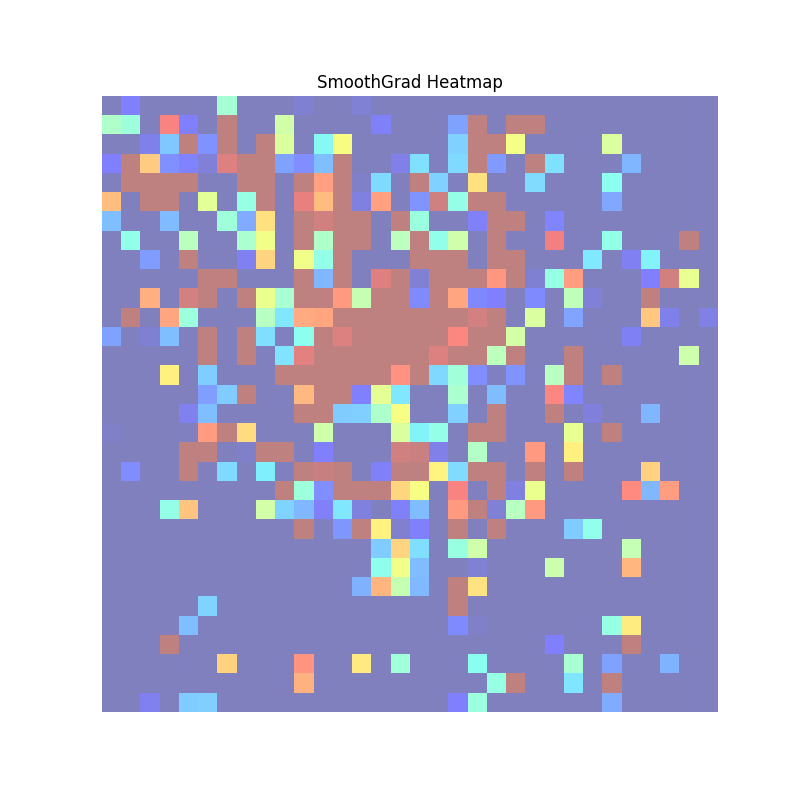}
        \caption{Smooth Grad}
    \end{subfigure}
    \begin{subfigure}{0.162\linewidth}
        \includegraphics[width=\linewidth]{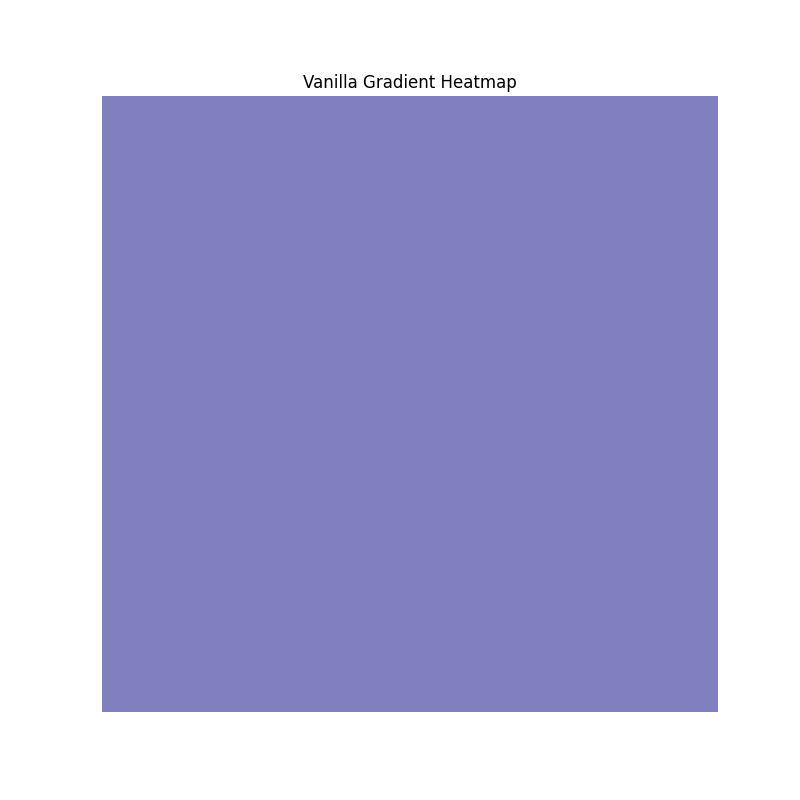}
        \caption{Vanilla Gradient}
    \end{subfigure}
    \begin{subfigure}{0.155\linewidth}
        \includegraphics[width=\linewidth]{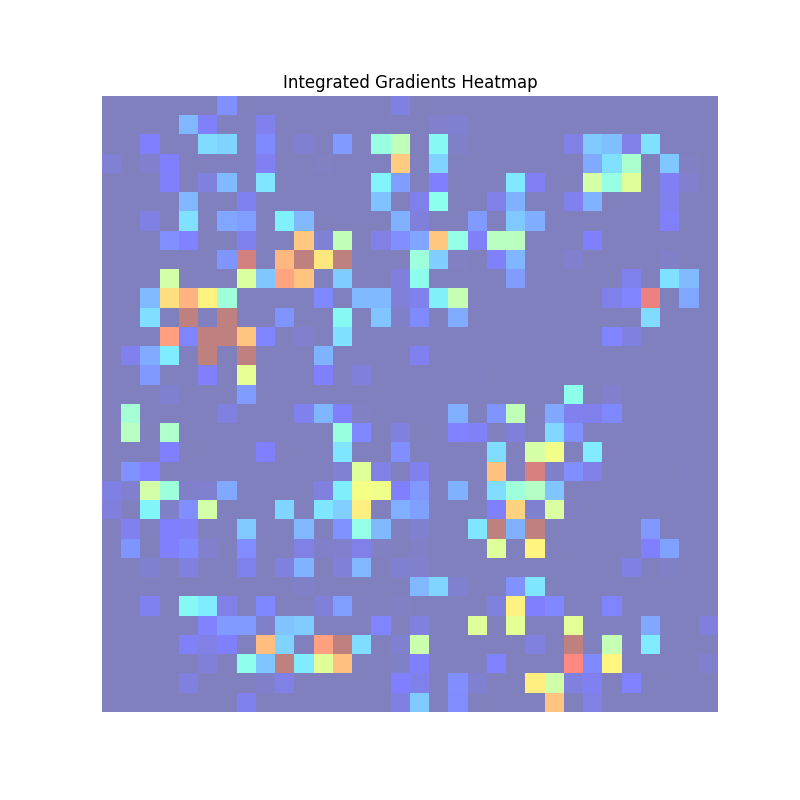}
        \caption{IG}
    \end{subfigure}
    \begin{subfigure}{0.162\linewidth}
        \includegraphics[width=\linewidth]{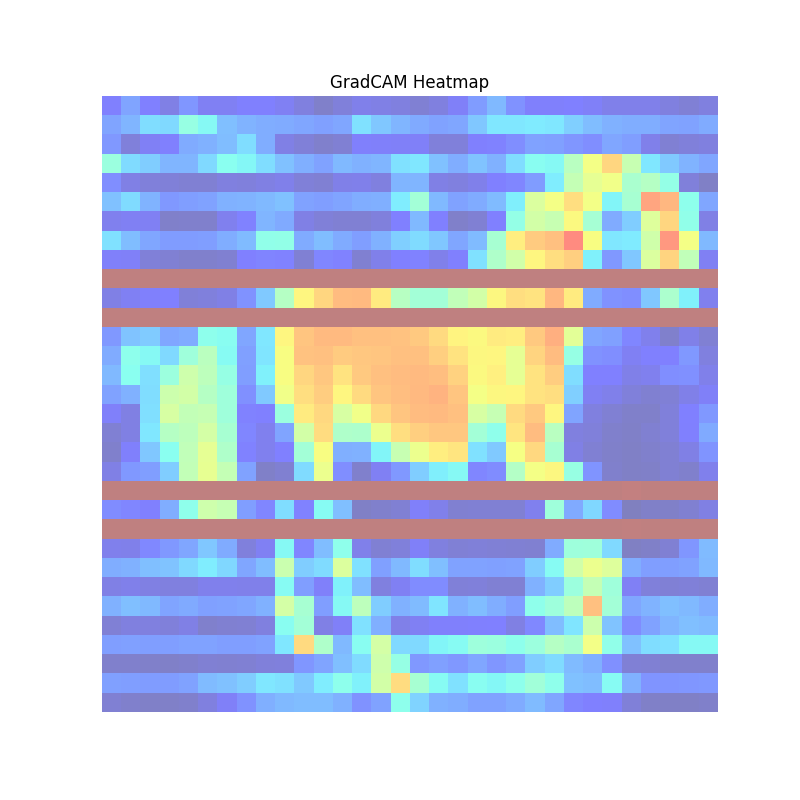}
        \caption{GradCAM}
    \end{subfigure}
    \caption{Visualizing ResNet's decisions on a Horse image of CIFAR10 Dataset using various explanation methods.}
    \label{fig: Multi-class Classification Interpretations-1}
\end{figure}

\subsubsection{Image Modality : Object Segmentation}
\begin{figure}[H]
    \centering
    \begin{subfigure}{0.23\linewidth}
        \centering
        \includegraphics[width=\linewidth]{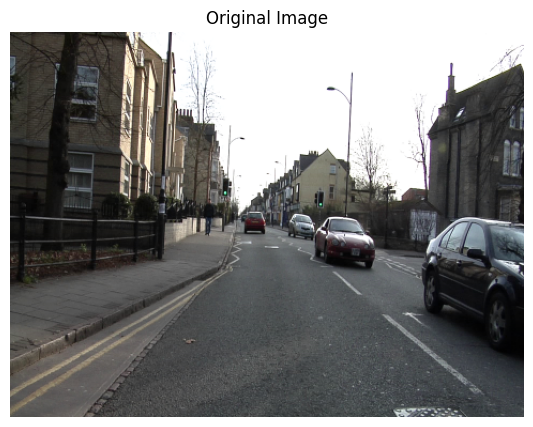}
        \caption{Original Image}
    \end{subfigure}
    \hspace{0.5em}
    \begin{subfigure}{0.23\linewidth}
        \centering
        \includegraphics[width=\linewidth]{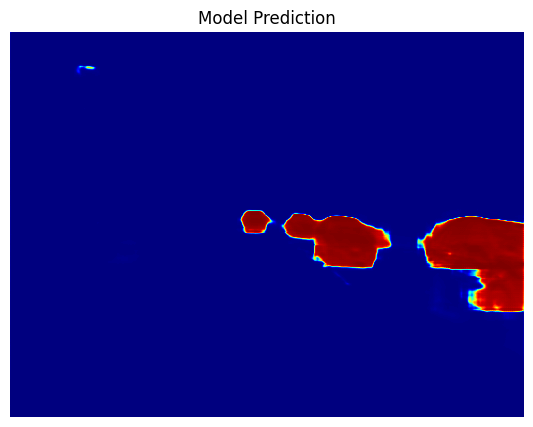}
        \caption{Model Prediction}
    \end{subfigure}
    \hspace{0.5em}
    \begin{subfigure}{0.23\linewidth}
        \centering
        \includegraphics[width=\linewidth]{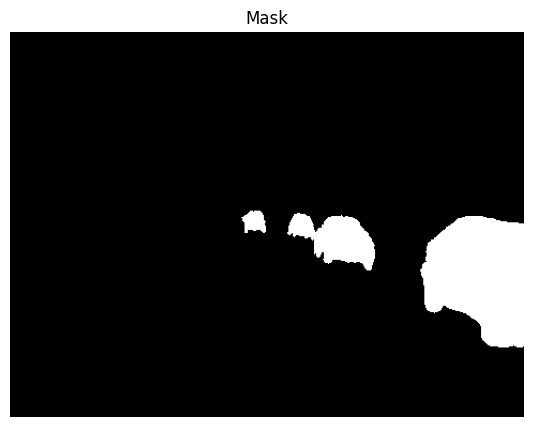}
        \caption{Label}
    \end{subfigure}
    \hspace{0.5em}
    \begin{subfigure}{0.23\linewidth}
        \centering
        \includegraphics[width=\linewidth]{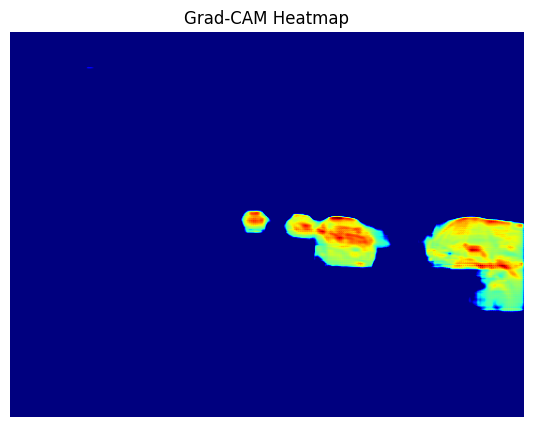}
        \caption{GradCAM}
    \end{subfigure}
    
    \vspace{0.5em} 

    \begin{subfigure}{0.24\linewidth}
        \centering
        \includegraphics[width=\linewidth]{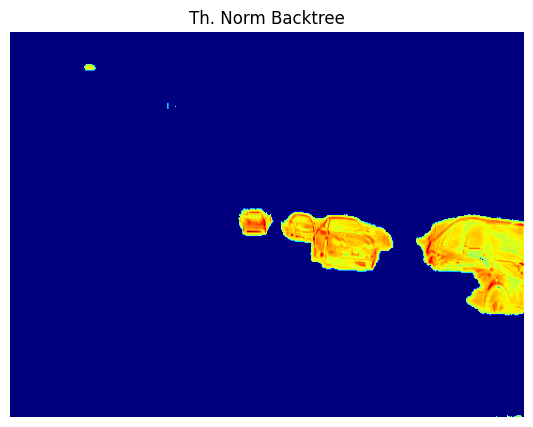}
        \caption{Backtrace Default}
    \end{subfigure}
    \hspace{0.5em}
    \begin{subfigure}{0.24\linewidth}
        \centering
        \includegraphics[width=\linewidth]{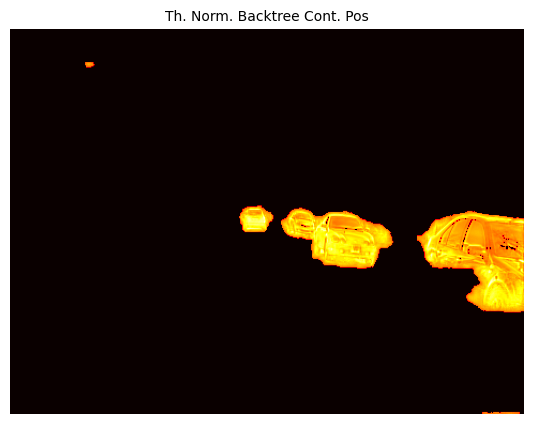}
        \caption{Backtrace Contrastive (Pos)}
    \end{subfigure}
    \hspace{0.5em}
    \begin{subfigure}{0.24\linewidth}
        \centering
        \includegraphics[width=\linewidth]{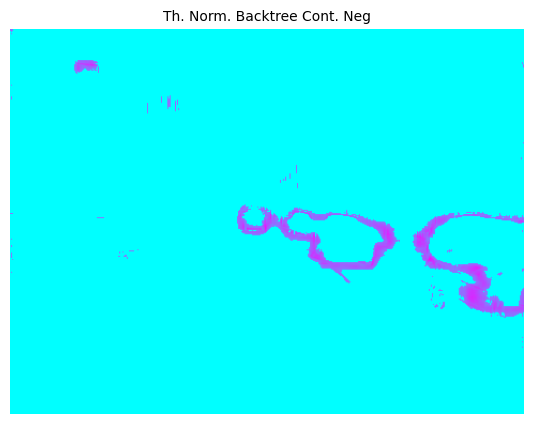}
        \caption{Backtrace Contrastive (Neg)}
    \end{subfigure}

    \caption{Analysis of a U-Net segmentation model's decision-making on a CamVid Dataset Sample. The figure shows the original Image, Model Prediction, and Label, alongside Explanations of GradCAM and Backtrace visualizations in Default and Contrastive modes.}
    \label{fig:segmentation-interpretations}
\end{figure}

\begin{figure}[H]
    \centering
    \begin{subfigure}{0.23\linewidth}
        \includegraphics[width=\linewidth]{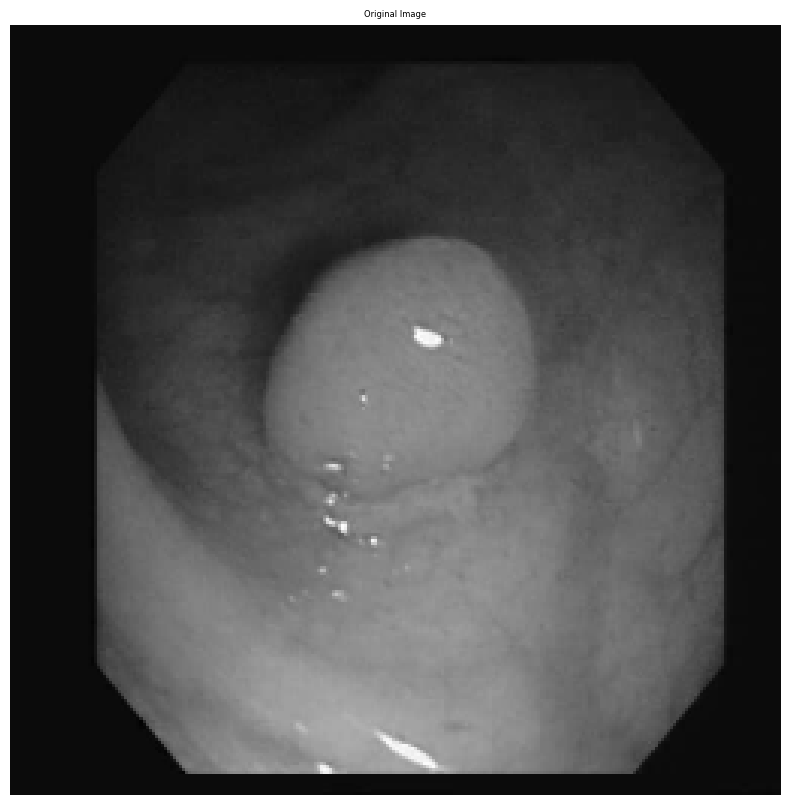}
        \caption{Original Image}
    \end{subfigure}
    \begin{subfigure}{0.23\linewidth}
        \includegraphics[width=\linewidth]{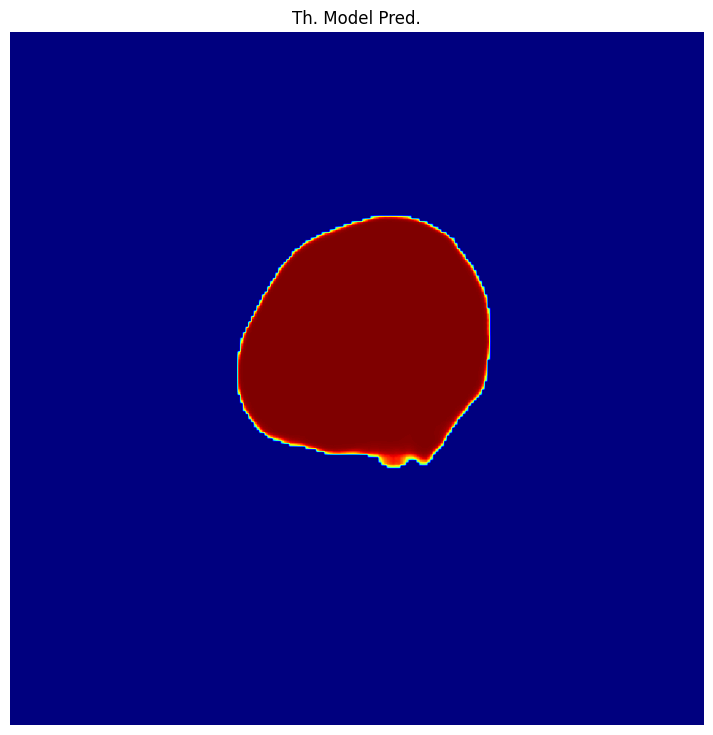}
        \caption{Prediction}
    \end{subfigure}
    \begin{subfigure}{0.23\linewidth}
        \includegraphics[width=\linewidth]{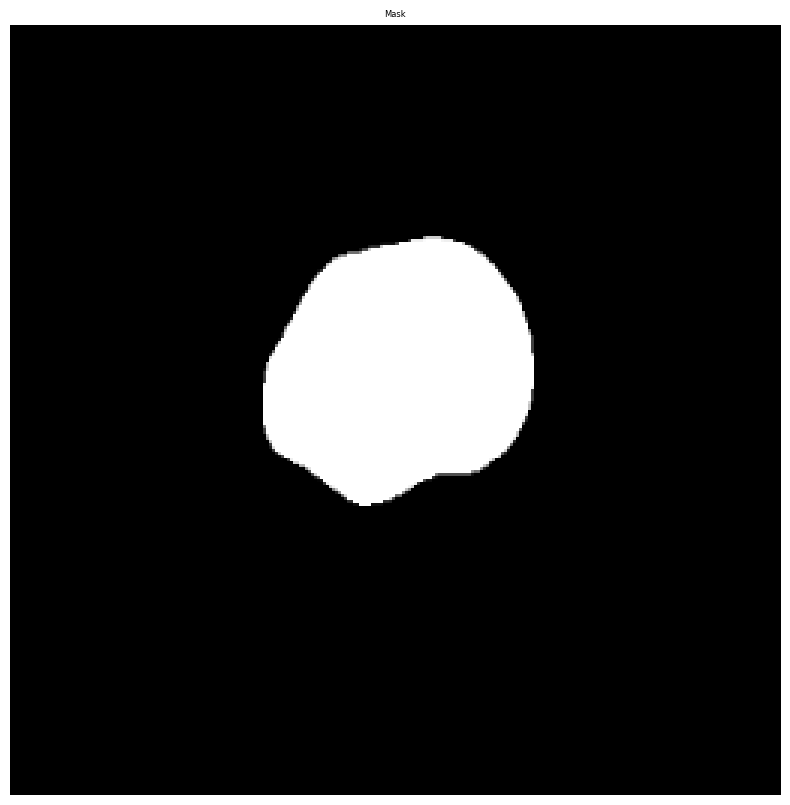}
        \caption{Label}
    \end{subfigure}
    \begin{subfigure}{0.23\linewidth}
        \includegraphics[width=\linewidth]{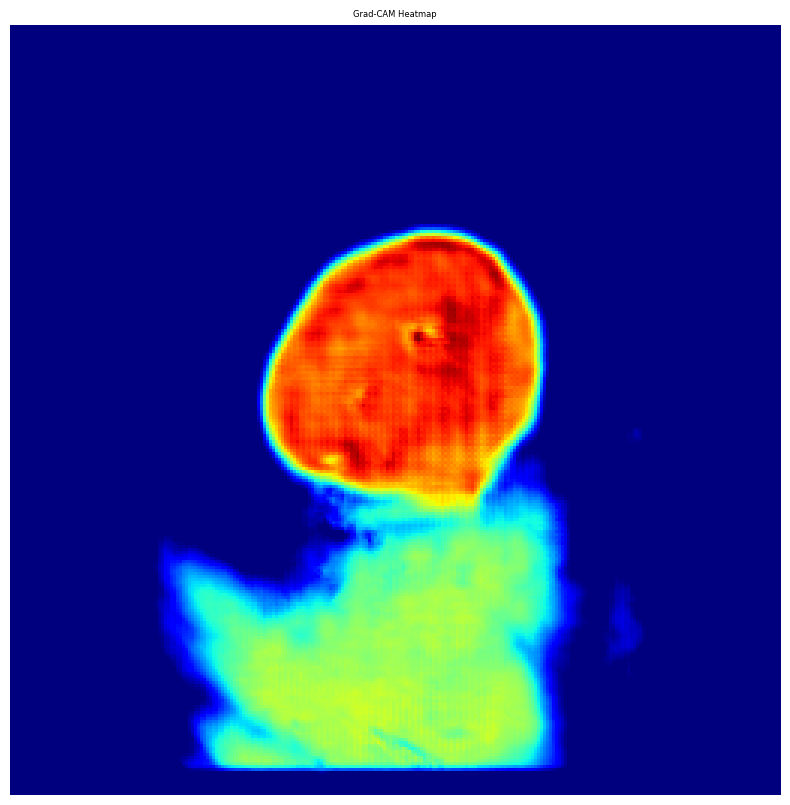}
        \caption{\parbox[t]{2cm}{GradCAM}}
    \end{subfigure}
    \hspace{0.5em}
    \begin{subfigure}{0.24\linewidth}
        \includegraphics[width=\linewidth]{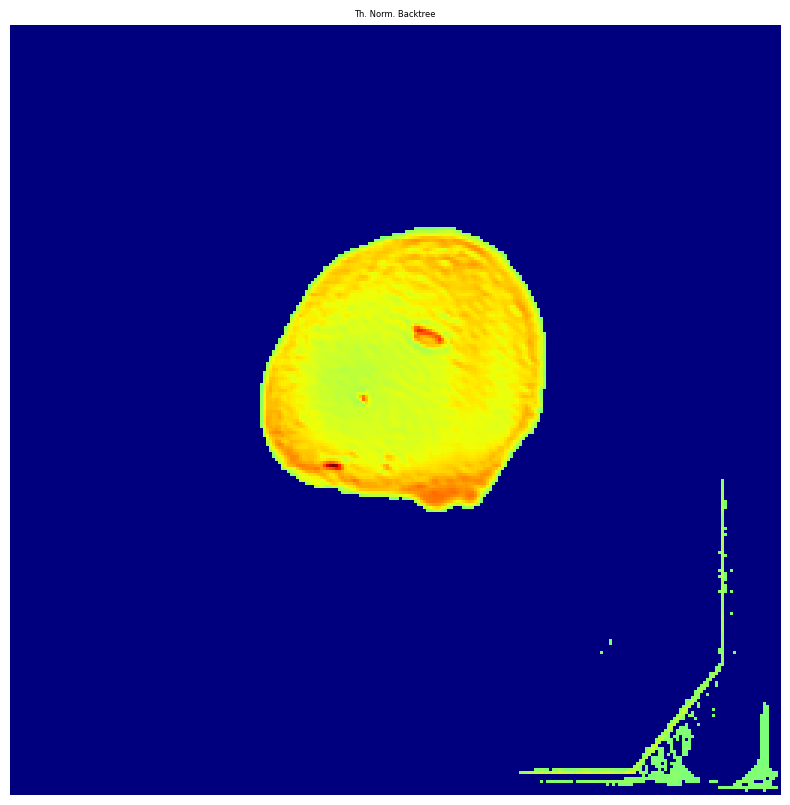}
        \caption{Backtrace Default}
    \end{subfigure}
    \begin{subfigure}{0.24\linewidth}
        \includegraphics[width=\linewidth]{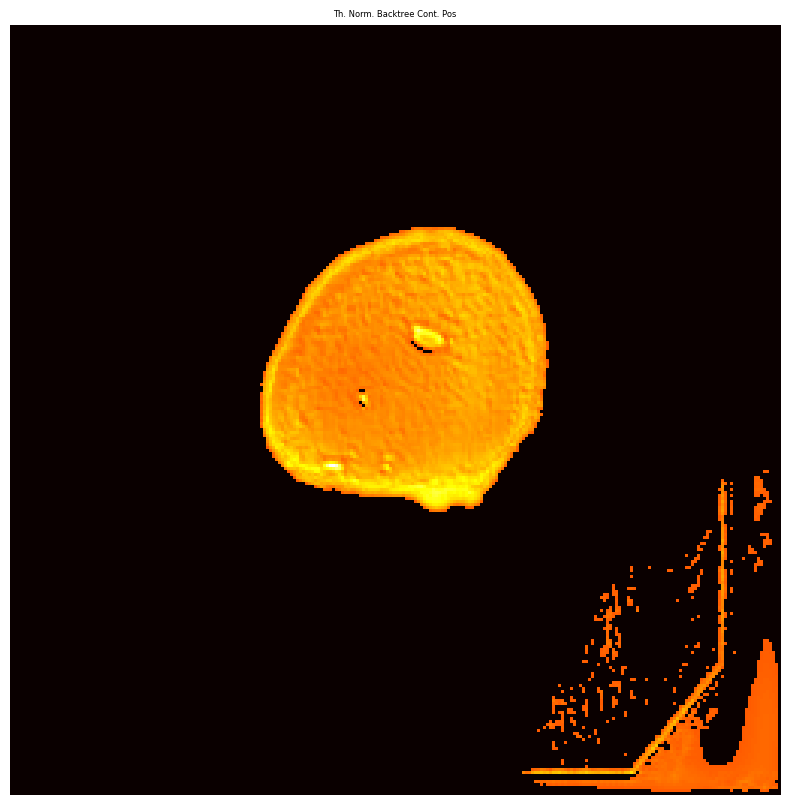}
        \caption{Backtrace Contrastive (Pos)}
    \end{subfigure}
    \begin{subfigure}{0.24\linewidth}
        \includegraphics[width=\linewidth]{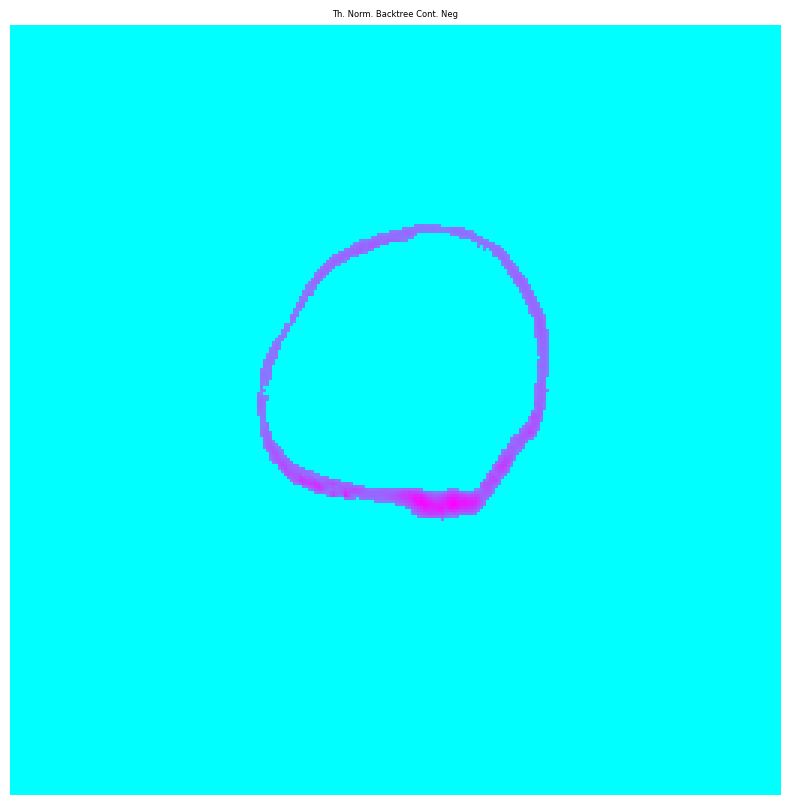}
        \caption{Backtrace Contrastive (Neg)}
    \end{subfigure}
    \caption{Analysis of a Tumour Segmentation Model's decision-making on a ClinicdB Dataset Sample. The figure shows the original Image, Model Prediction, and Label, alongside Explanations of GradCAM and Backtrace visualizations in Default and Contrastive modes.}
    \label{fig: Segmentation Interpretations-2}
\end{figure}

\subsubsection{Image Modality : Object Detection}

\begin{figure}[H]
    \centering
    \begin{subfigure}{0.20\linewidth}
        \includegraphics[width=\linewidth]{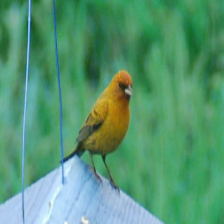}
        \caption{Original Image}
    \end{subfigure}
    \begin{subfigure}{0.20\linewidth}
        \includegraphics[width=\linewidth]{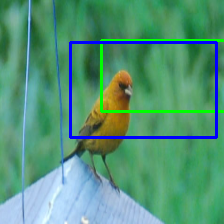}
        \caption{Prediction $\And$ Label}
    \end{subfigure}
    \begin{subfigure}{0.20\linewidth}
        \includegraphics[width=\linewidth]{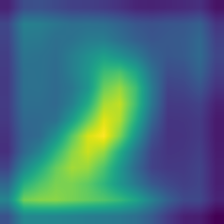}
        \caption{GradCAM}
    \end{subfigure}
    \begin{subfigure}{0.20\linewidth}
        \includegraphics[width=\linewidth]{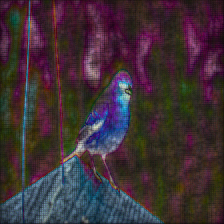}
        \caption{Backtrace}
    \end{subfigure}
    \caption{Explanations of the model's decision-making process on a Bird image from the CUB-200 dataset, using Grad-CAM and Backtrace to highlight the key regions influencing the prediction.}
    \label{fig: Object_Detection_Interpretations-1}
\end{figure}

\begin{figure}[H]
    \centering
    \begin{subfigure}{0.20\linewidth}
        \includegraphics[width=\linewidth]{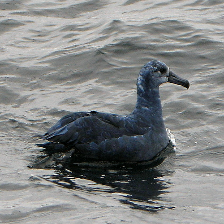}
        \caption{Original Image}
    \end{subfigure}
    \begin{subfigure}{0.20\linewidth}
        \includegraphics[width=\linewidth]{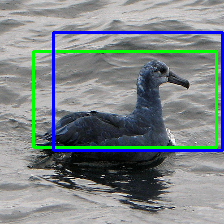}
        \caption{Prediction $\And$ Label}
    \end{subfigure}
    \begin{subfigure}{0.20\linewidth}
        \includegraphics[width=\linewidth]{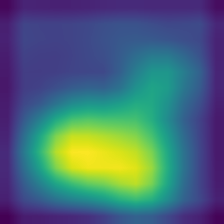}
        \caption{GradCAM}
    \end{subfigure}
    \begin{subfigure}{0.20\linewidth}
        \includegraphics[width=\linewidth]{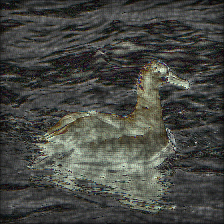}
        \caption{Backtrace}
    \end{subfigure}
    \caption{Explanations of the model's decision-making process on a duck image from the CUB-200 dataset, using Grad-CAM and Backtrace to highlight the key regions influencing the prediction.}
    \label{fig:Object_Detection_Interpretations-2}
\end{figure}

\subsubsection{Text Modality : BERT Sentiment Classification}
\label{sec:text-samples}
\begin{figure}[H]
    \centering
    \includegraphics[width=0.8\textwidth]{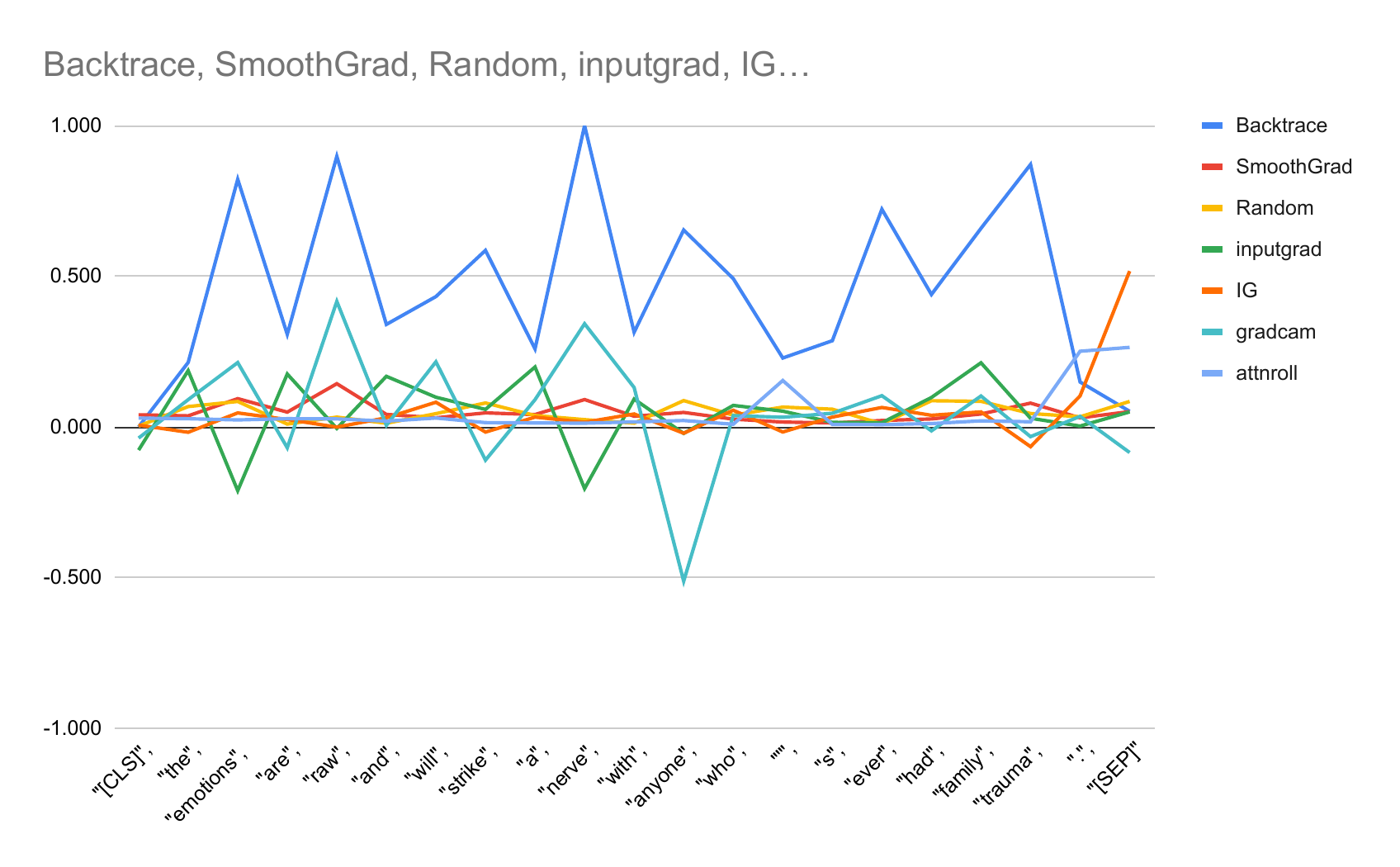}
    \caption{Explanations by different methods for model decision making for Sentiment Analysis for a sample from SST Dataset.  \textbf{Input Text:} The emotions are raw and will strike a nerve with anyone who ever had family trauma. \textbf{Prediction: 1} and \textbf{Label: 1}}
    \label{fig:bert_text}
\end{figure}
\nopagebreak 

\subsubsection{Text Modality : Multi-Class Classification}
\begin{figure}[H]
    \centering
    \begin{subfigure}{0.42\linewidth}
        \includegraphics[width=\linewidth]{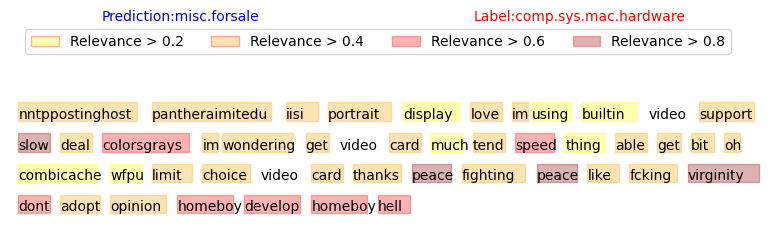}
        \caption{Original Image}
    \end{subfigure}
    \begin{subfigure}{0.42\linewidth}
        \includegraphics[width=\linewidth]{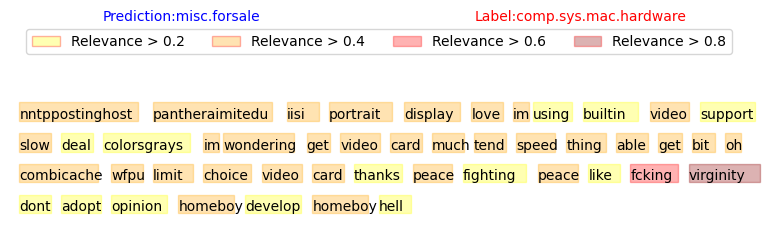}
        \caption{Prediction $\And$ Label}
    \end{subfigure}
    \begin{subfigure}{0.85\linewidth}
        \includegraphics[width=\linewidth]{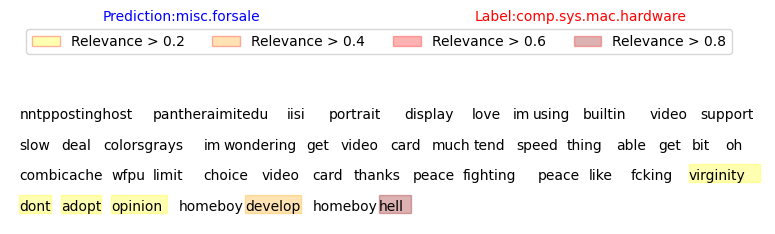}
        \caption{Backtrace}
    \end{subfigure}
    \caption{Explanations of the model's decision-making process for multi-class topic detection for incorrect classification on a Model using Pre-Trained Glove Word Embedding and 1D CNN, using LIME, SHAP and Backtrace to highlight the key regions influencing the prediction.}
    \label{fig: Word-Embedding-Topic-MC-2}
\end{figure}

\begin{figure}[H]
    \centering
    \begin{subfigure}{0.42\linewidth}
        \includegraphics[width=\linewidth]{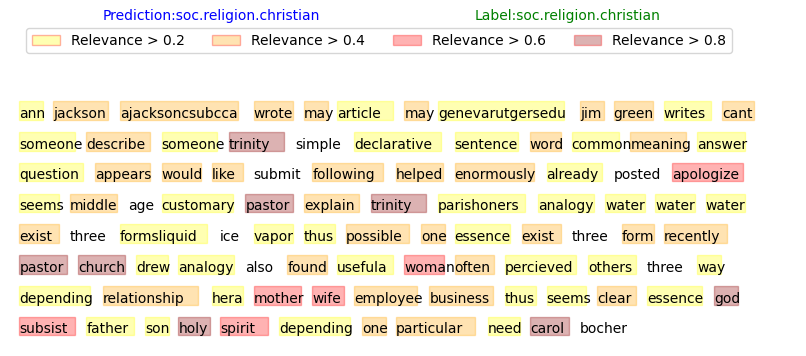}
        \caption{LIME}
    \end{subfigure}
    \begin{subfigure}{0.42\linewidth}
        \includegraphics[width=\linewidth]{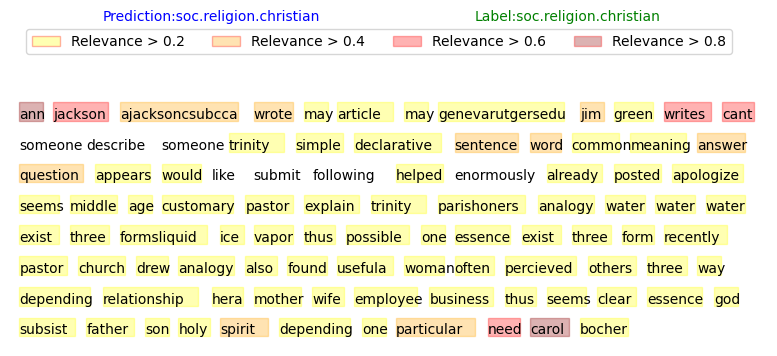}
        \caption{SHAP}
    \end{subfigure}
    \begin{subfigure}{0.8\linewidth}
        \includegraphics[width=\linewidth]{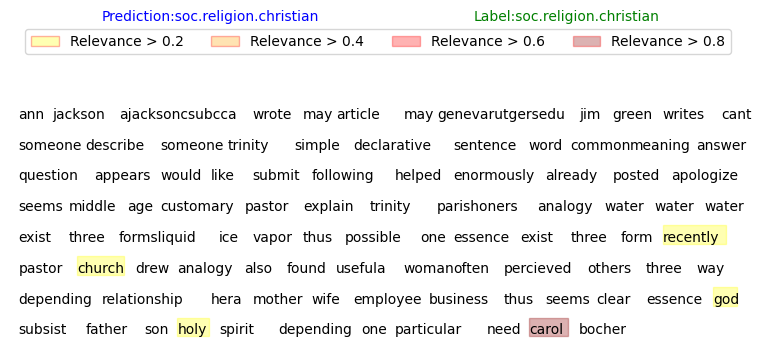}
        \caption{Backtrace}
    \end{subfigure}
    \caption{Explanations of the model's decision-making process for multi-class topic detection for correct classification on a Model using Pre-Trained Glove Word Embedding and 1D CNN, using LIME, SHAP and Backtrace to highlight the key regions influencing the prediction.}
    \label{fig: Word-Embedding-Topic-MC-1}
\end{figure}


\subsubsection{Text Modality: Llama3.2}
\begin{figure}[H] 
    \centering
    \includegraphics[width=\textwidth]{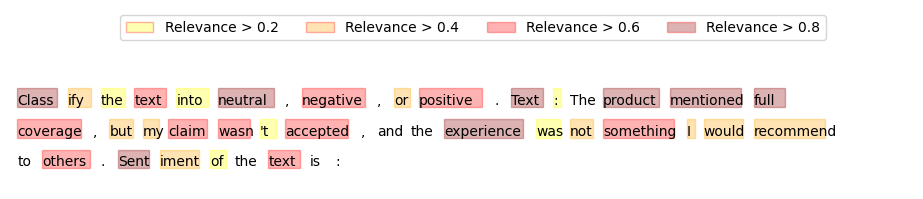} 
    \caption{\\\textbf{Input Prompt:} Classify the text into neutral, negative, or positive. Text: The product mentioned full coverage, but my claim wasn't accepted, and the experience was not something I would recommend to others. Sentiment of the text is:\\
    \textbf{Ouput : } Negative. Reason: The text expresses dissatisfaction with the product and the experience, stating that the claim was not accepted and the experience was not something the author would recommend. This indicates a negative sentiment. Note: The text does not contain any explicit negative
}
    \label{fig:prompt1} 
\end{figure}

\begin{figure}[H] 
    \centering
    \includegraphics[width=\textwidth]{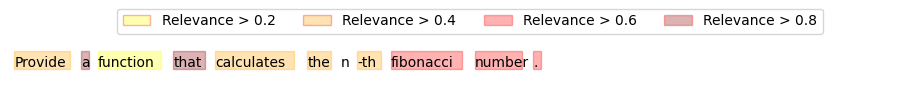} 
        \caption{\textbf{Input Prompt:} Provide a function that calculates the n-th fibonacci number.
        \\
        \textbf{Output: } 
        Fibonacci sequence is a series of numbers in which each number is the sum of the two preceding ones, usually starting with 0 and 1.Fibonacci Function. Here's a Python function that calculates the n-th Fibonacci number using memoization to improve performance.
        \\
        def fibonacci(n, memo={}):
        \\Calculate the n-th Fibonacci number.
        \\Args:
        n (int): The position of the Fibonacci number to calculate.
        memo (dict): A dictionary to store previously calculated Fibonacci numbers.
        \\Returns:
        int: The n-th Fibonacci number.
        \\if n <= 0:
        \\    return 0
        \\elif n == 1:
        \\    return 1
        \\elif n not in memo:
        \\    memo[n] = fibonacci(n-1, memo) + fibonacci(n-2, memo)
        \\return memo[n]
    \\Example usage:
    print(fibonacci(10))  
    \\Output: 55}
    \label{fig:prompt5} 
\end{figure}

\begin{figure}[H] 
    \centering
    \includegraphics[width=\textwidth]{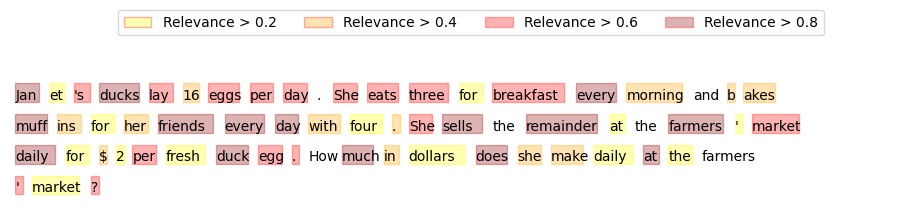} 
    \caption{\\\textbf{Input Prompt:} Janet's ducks lay 16 eggs per day. She eats three for breakfast every morning and bakes muffins for her friends every day with four. She sells the remainder at the farmers' market daily for \$2 per fresh duck egg. How much in dollars does she make daily at the farmers' market?
    \\\textbf{Output: } Step 1: Calculate the number of eggs laid per day Janet's ducks lay 16 eggs per day. Step 2: Calculate the number of eggs eaten per day. Janet eats 3 eggs for breakfast every morning.}
    \label{fig:prompt2} 
\end{figure}

\newpage

\bibliographystyle{unsrt}  
\bibliography{references}  

\end{document}